\documentclass[journal,onecolumn]{IEEEtran}
\usepackage{amssymb,amsmath}
\usepackage{multirow}
\usepackage{graphicx}
\usepackage{epstopdf}
\usepackage{wrapfig}
\newtheorem{theorem}{Theorem}[section]
\newtheorem{lemma}{Lemma}[section]

\newtheorem{remark}{Remark}[section]

\DeclareMathOperator*{\argmin}{arg\,min}
\def \dx{\, \mathrm{d}}

\begin{document}

\title{Clustering and Community Detection with Imbalanced Clusters}

\author{Cem Aksoylar, Jing Qian, Venkatesh Saligrama%
  \thanks{This work was supported in part by the U.S.\ Department of Homeland Security, Science and Technology Directorate, Office of University Programs, under Grant Award 2013-ST-061-ED0001, in part by the National Science Foundation under Grants CCF-1320547 and 1218992, and in part by ONR Grant 50202168.}%
  \thanks{Cem Aksoylar and Venkatesh Saligrama are with the Department of Electrical and Computer Engineering at Boston University, Boston, MA. Jing Qian is with Facebook, Menlo Park, CA.}}

\IEEEpubid{Copyright~\copyright~2016 IEEE.}

\maketitle

\begin{abstract}
  Spectral clustering methods which are frequently used in clustering and community detection applications are sensitive to the specific graph constructions particularly when imbalanced clusters are present.
  We show that ratio cut (RCut) or normalized cut (NCut) objectives are not tailored to imbalanced cluster sizes since they tend to emphasize cut sizes over cut values.
  We propose a graph partitioning problem that seeks minimum cut partitions under minimum size constraints on partitions to deal with imbalanced cluster sizes. Our approach parameterizes a family of graphs by adaptively modulating node degrees on a fixed node set, yielding a set of parameter dependent cuts reflecting varying levels of imbalance. The solution to our problem is then obtained by optimizing over these parameters. We present rigorous limit cut analysis results to justify our approach and demonstrate the superiority of our method through experiments on synthetic and real datasets for data clustering, semi-supervised learning and community detection.
\end{abstract}

\section{Introduction}\label{sec:intro_motiv}
We consider graph partitioning problems with imbalanced partition sizes for two different graph modalities: similarity networks where we are given a measure of similarity for each pair of nodes (such as distance) and connectivity networks where we have a set of nodes and unweighted edges between pairs of nodes. The first modality is used frequently for graph-based spectral methods for clustering and semi-supervised learning (SSL) tasks. In this context, data with imbalanced clusters arises in many learning applications and has attracted much interest \cite{HeGarcia09}. The second modality is the setup for community detection problems where identification of communities with small sizes has been considered in the literature \cite{shah}. 

In spectral methods for clustering and SSL, first a graph representing the data is constructed and then spectral clustering (SC) \cite{Hagen92,Shi00} or SSL algorithms \cite{Zhu08,WanJebCha08} are applied on the resulting graph. Common graph construction methods include $\epsilon$-graphs, fully-connected RBF-weighted (full-RBF) graphs and $k$-nearest neighbor ($k$-NN) graphs. Of the three, $k$-NN construction appears to be most popular due to its relative robustness to outliers \cite{Zhu08,Luxburg07}.
Recently \cite{JebShc06} proposed $b$-matching graphs which claim to eliminate some of the spurious edges of $k$-NN graphs and lead to better performance.
Model-based approaches that incorporate imbalancedness have previously been investigated \cite{Fraley02}, however they typically assume simple cluster shapes and need multiple restarts. In contrast non-parametric graph-based approaches do not have this issue and are able to capture complex shapes \cite{Ng01}. 

Graph partitioning methods on connectivity networks based on spectral clustering are frequently used in the literature \cite{white} for the problem of community detection \cite{community}. One limitation of spectral methods is that they often fail to detect smaller community structures in dense networks \cite{shah}. While this limitation can be observed empirically, there also exist recent theoretical results for the stochastic block model \cite{hero,nadakuditi} that allows the quantification of the difficulty of detection relative to community sizes. 

We also remark that community detection problems are not limited to the connectivity network setup and the similarity network modality is also useful when there exists additional information associated with the nodes or edges. One example of such a problem is a citation network, where each node represents an academic paper with text content. Similarity between papers can be quantified by extracting topics from the documents and using the topic distributions in two papers to compute a similarity score. This information can be used in addition to citation information (which by itself is a connectivity network) to discover communities, e.g.~corresponding to research fields. These kinds of approaches combining the two modalities have been investigated in the literature \cite{yang,topiclink}.

To the best of our knowledge, systematic ways of adapting spectral methods to imbalanced data do not exist. We show that the poor performance of spectral methods on imbalanced data can be attributed to applying ratio cut (RCut) or normalized cut (NCut) minimization objectives on traditional graphs, which sometimes tend to emphasize balanced partition size over small cut-values.

\begin{figure}[tbp]
  \centering
  \includegraphics[width=0.45\textwidth]{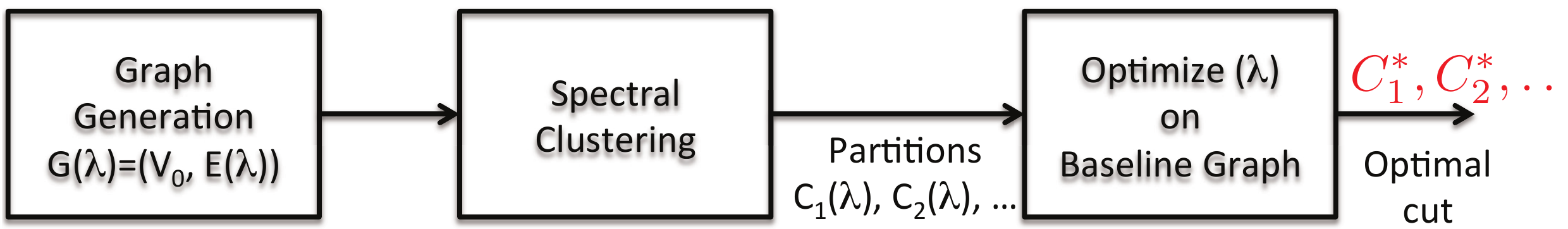}
  \caption{\small Proposed framework for clustering on imbalanced data.}
  \label{f.framework}
\end{figure}

\vspace{5pt}
\noindent
{\bf Our contributions:} \\
To deal with imbalanced data we propose the partition constrained minimum cut problem (PCut). We remark that size-constrained min-cut problems appear to be computationally intractable \cite{Galbiati11,Ji04}, thus instead we attempt to solve PCut on a parameterized family of cuts. To realize these cuts we parameterize a family of graphs over some parametric space $\lambda \in \Lambda$ and generate candidate cuts using spectral methods as a black-box. This requires a sufficiently rich graph parameterization capable of approximating varying degrees of imbalanced data. To this end we introduce a novel parameterization for graphs that involves adaptively modulating node degrees in varying proportions, for both similarity and connectivity networks. We then solve PCut on a baseline graph over the candidate cuts generated using this parameterization. Fig.~\ref{f.framework} depicts our approach for binary clustering. Our limit cut analysis shows that our approach asymptotically does adapt to imbalanced and proximal clusters. We then demonstrate the superiority of our method through unsupervised clustering, semi-supervised learning and community detection experiments on synthetic and real datasets.
Note that we do not presume imbalancedness in the underlying cluster sizes. Our method significantly outperforms traditional approaches when the underlying clusters are imbalanced, while remaining competitive when they are balanced. Our paper is based in part on preliminary results described in \cite{qiansaligrama14}.

\noindent
{\bf Related work:} \\
Sensitivity of spectral methods to graph construction in similarity networks is well documented \cite{Luxburg07,Maier1,JebWanCha09}. \cite{Zelnik04} suggests an adaptive RBF parameter in full-RBF graphs to deal with imbalanced clusters. \cite{Nadler07} describes these drawbacks from a random walk perspective.
\cite{Buhler09,ShiBelkinYu09} also mention imbalanced clusters, but none of these works explicitly deal with imbalanced data. We remark that our approach is complementary to their schemes and can be used in conjunction.
Another related approach is size-constrained clustering \cite{ST97,Feige03,Andreev04,Hoppner08,ZhuWangLi10,Galbiati11}, which is shown to be NP-hard. \cite{Nagano11} proposes submodularity based schemes that work only for certain special cases.
In addition, these works either impose exact cardinality constraints or upper bounds on the cluster sizes to look for balanced partitions.
While this is related, we seek minimum cuts with lower bounds on smallest-sized clusters.
Minimum cuts with lower bounds on cluster size naturally arises because we seek cuts at density valleys (accounted for by the min-cut objective) while rejecting singleton clusters and outliers (accounted for by cluster size constraint).
It is not hard to see that our problem is computationally no better than min-cut with upper bounds of size constraints.\footnote{In the 2-way partition setting, min-cut with lower bounds is equivalent to min-cut with upper bounds and is thus NP-hard. The multi-way partition problem generalizes the 2-way setting. }

A related area of recent active research is the detection of anomalous clusters in signals over networks \cite{aistats14,nips14,knn-anomaly}. This line of work focuses on the \emph{detection} of well-connected subgraphs of given network data, which complements the approach considered here which aims to \emph{partition} the given graph to well-connected subgraphs.

The organization of the paper is as follows.
In Sec.~\ref{sec:motiv} we propose our partition constrained min-cut (PCut) framework, illustrate some of the fundamental issues underlying poor performance of spectral methods on imbalanced data and explain how PCut can deal with it.
We describe the details of our PCut algorithm for both similarity and connectivity networks in Sec.~\ref{sec:RMD_idea}, and explore its theoretical basis for the similarity network framework in Sec.~\ref{sec:thm}.
In Sec.~\ref{sec:experiment} we present experiments on synthetic and real datasets to show significant improvements in SC, SSL and community detection tasks.
Sec.~\ref{sec:conclusion} concludes the paper.

\section{Partition Constrained Min-Cut (PCut)}\label{sec:motiv}

We formalize the PCut problem for the similarity network and connectivity network modalities. 
Let $G=(V,\,E,\,W)$ be a weighted undirected graph with $n$ nodes, where the weights $(W)_{u,v} = w(u,v)$ are similarity measures between two nodes, which is equal to 1 uniformly for connectivity graphs. 
We denote by $S$ a cut that partitions $V$ into $C_S$ and $\bar{C}_S$. The cut-value associated with $S$ is:
\begin{equation}\label{equ:cut}
  Cut(C_S,\bar{C}_S) = \sum_{u\in C_S, v\in \bar{C}_S, (u,v)\in E} w(u,v).
\end{equation}
For similarity graphs the edge set $E$ may be generated from the similarity weights $W$, by constructions such as the $k$-NN graph where each node is connected to its $k$-closest (i.e.~most similar) neighbors, $\epsilon$-graph where nodes are connected to all other nodes with a larger than $\epsilon$ similarity value, or a full-RBF graph with edges between all node pairs. We refer to these methods for edge set generation as graph construction methods.

We pose the problem of partition size constrained minimum cut (PCut) as:
\begin{equation} \label{e.empopt}
  S^* = \argmin_{S} \left \{ Cut(C_S,\bar{C}_S) \mid \min\{|C_S|,\,|\bar{C}_S|\} \geq \delta |V| \right \}, \\ 
\end{equation}
where we also denote the partition of nodes corresponding to the optimal cut $S^*$ with $(C^*,\bar{C}^*)$.

While our proposed method does not necessitate features associated with nodes in similarity networks and requires only that similarity scores exist for each pair of nodes, for analysis purposes we consider a generative model where we assume that each node $v$ has features $x_v \in \mathbb{R}^d$ that is drawn from some unknown density $f(x_v)$. In this generative framework, PCut corresponds to searching for a hypersurface $S$ that partitions $\mathbb{R}^d$ into two subsets $D$ and $\bar D$ (with $D \cup \bar D=\mathbb{R}^d$) with non-trivial mass and passes through low-density regions.

We remark that while such a formulation is natural for learning problems such as clustering or SSL, such representations arise in many other problems where we have additional information associated with nodes or edges. One example is a collaborative filtering problem for recommender systems with users where user ratings for movies can be represented with a linear factor model $R = UV$, where each user is represented by a latent feature vector $U_i$ and each movie by $V_j$. Then probabilistic matrix factorization methods such as \cite{bpmf} induce a probability distribution over users' latent features $u \sim f(u)$ that are independent and identically distributed. 
Another example is citation networks, where each node represents an academic paper with text content. In this case each node can be represented by a distribution $\theta \sim f(\theta)$ over topics using topic modeling, which are sampled from induced probability spaces in generative methods such as latent Dirichlet allocation \cite{lda}. 

Throughout the paper, we refer to $\alpha = \frac{\min\{n_1,n_2\}}{n} \leq \frac{1}{2}$ as the imbalance parameter for a graph with two underlying clusters with sizes $n_1$ and $n_2 = n - n_1$. Note that for the generative model this corresponds to $\alpha = \min\{\mu(D),\mu(\bar D)\}$ where $(D, \bar D)$ is the optimal size-constrained partitioning.

Eq.~\eqref{e.empopt} describes a binary partitioning problem but generalizes to arbitrary number of partitions, for which we state PCut as: 
\begin{align*}
  (C_1^*, \ldots,C_K^*) = \argmin_{C_1,\ldots,C_K} \sum_{i=1}^K Cut(C_i,\bar{C}_i) \mathrm{~~s.t.~~} \min_i |C_i| \geq \delta |V|,\; \bigcup_i C_i = V,\; C_i \cap C_j = \emptyset \; \forall i, j.
\end{align*}
While we consider the binary problem with the hypersurface interpretation of cuts $S$ for analyzing limit-cut behavior, we present our algorithm in Sec.~\ref{sec:RMD_idea} for the generalized problem and utilize it in experiments in Sec.~\ref{sec:experiment}.

Note that without size constraints the problem in Eq.~\eqref{e.empopt} is identical to the min-cut criterion \cite{Stoer97}, which is well-known to be sensitive to outliers. This objective is closely related to the problem of graph partitioning with size constraints, with various versions known to be NP-hard~\cite{Ji04}. Approximations to such partitioning problems have been developed~\cite{Andreev04} but appear to be overly conservative. More importantly these papers~\cite{Andreev04,Hoppner08,ZhuWangLi10} either focus on balanced partitions or cuts with exact size constraints. In contrast our objective here is to identify natural low-density cuts that are not too small (i.e.~with lower bounds on the smallest sized cluster).
We employ SC as a black-box to generate candidate cuts on a suitably parameterized family of graphs. Eq.~\eqref{e.empopt} is then optimized over these candidate cuts.

\subsection{RCut, NCut and PCut}\label{subsec:alg}
The well-known spectral clustering algorithms attempt to minimize RCut or NCut:
\begin{equation}\label{equ:ratiocut}
  \min_S\,\,Cut(C_S,\bar{C}_S) \left( \frac{size(V)}{size(C_S)} + \frac{size(V)}{size(\bar{C}_S)} \right),
 \end{equation}
where $size(C)=|C|$ for RCut and $size(C)=\sum_{u\in C,v\in V}w(u,v)$ for NCut, and for simplicity we considered the binary problem. Both objectives seek to trade-off low cut-values against cut size.
While robust to outliers, minimizing RCut/NCut can lead to poor performance when data is imbalanced (i.e. with small $\alpha$).
To see this, we define cut-ratio $q \in [0,\,1]$ and imbalance coefficient $y \in [0,\,0.5]$ for some graph $G=(V,E,W)$
\[ q = {Cut(C^*,\bar{C}^*) \over Cut(C_B,\bar{C}_B)},\,\,\,y= \frac{\min \{size(C^*),\,size(\bar{C}^*)\}}{size(C^*)+size(\bar{C}^*)}, \]
where $(C^*,\bar C^*)$ corresponds to optimal PCut and $S_B(C_B,\bar C_B)$ is any balanced partition with $size(C_B)=size(\bar C_B)$. 

\begin{figure}[htbp]
  \centering
  \includegraphics[width=0.5\textwidth]{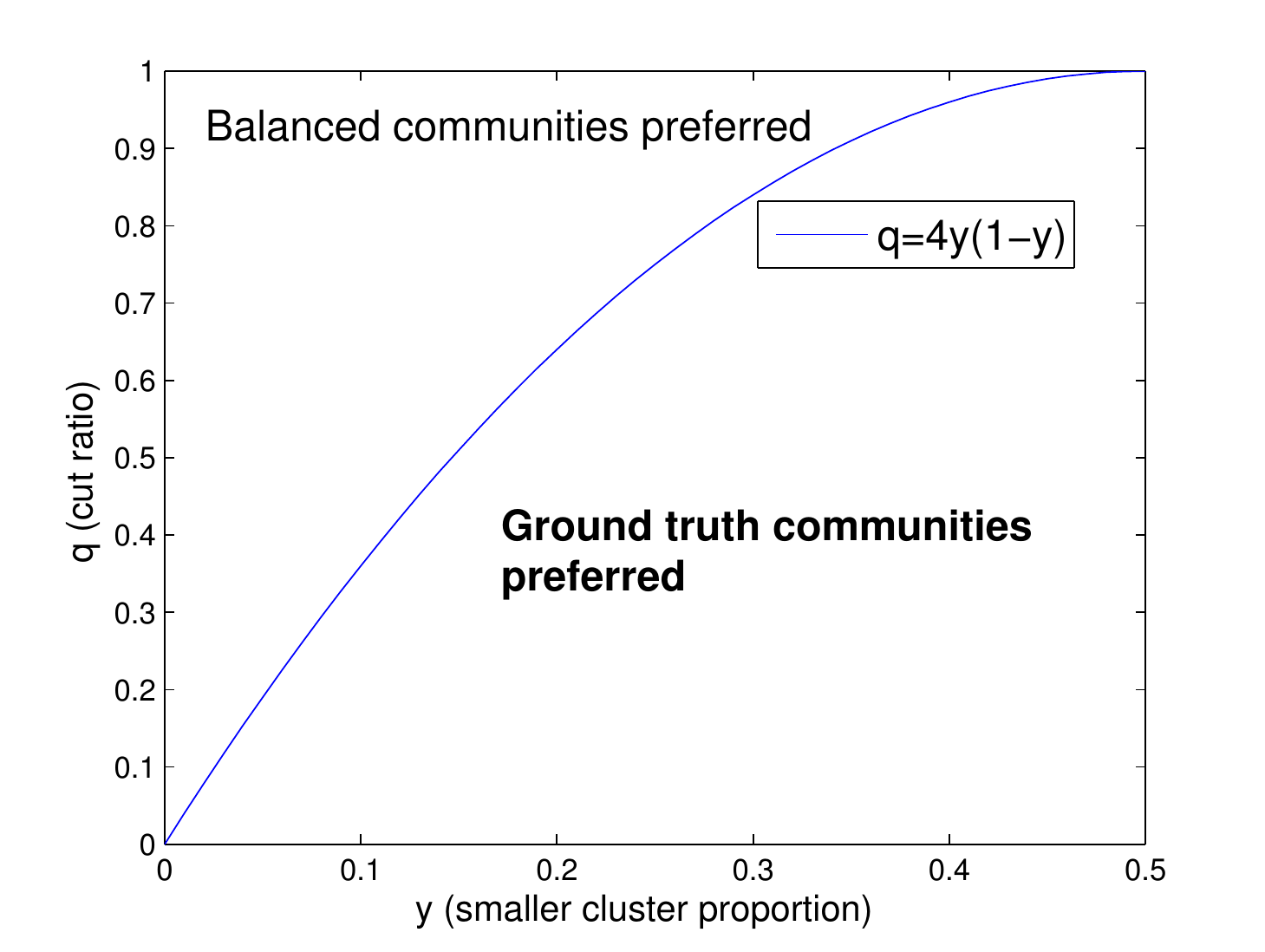}
  \caption{\small Cut-ratio ($q$) vs.\ imbalance ($y$). RCut value is smaller for balanced cuts than imbalanced optimal cuts for cut-ratios above the curve.}
  \label{fig:qy}
\end{figure}

\begin{remark}\label{remark}
  For any partition $(C, \bar C)$ 
  \[ RCut(C,\bar C) = Cut(C, \bar C) \left ({1 \over |C|} +  {1 \over |\bar C|} \right ) \implies {RCut(C^*,\bar C^*) \over RCut(C_B,\bar C_B)} = {q \over 4 y(1-y)}. \]
A similar expression holds for NCut with appropriate modifications. Because $q$ varies for different graphs but $y$ does not, the ratio $q/4y(1-y)$ depends on the underlying graph in a connectivity network and the specific graph construction in a similarity network.
So it is plausible that for some graphs RCut/NCut value satisfies $q > 4y(1-y)$ while for others $q < 4y(1-y)$. In the former case RCut/NCut will favor a balanced cut over the ground-truth cut $(C^*,\bar C^*)$ and vice versa if the latter is true. Fig.~\ref{fig:qy} illustrates this behavior.
\end{remark}

We analyze the limit-cut behavior of $k$-NN, $\epsilon$-graphs and RBF graphs to build intuition for the similarity network case. For properly chosen $k_n$, $\sigma_n$ and $\epsilon_n$ \cite{Maier1,Narayanan06}, as sample size $n\rightarrow \infty$, we have
\begin{equation} \label{e.lcutqy}
q \longrightarrow {\int_{S_0} f^{\gamma}(x) \dx x \over \int_{S_B} f^{\gamma}(x) \dx x},\,\,\,\,\, y \longrightarrow \min \{\mu(D_0),\,\mu(\bar D_0)\} = \alpha,
\end{equation}
where $\gamma<1$ for $k$-NN and $\gamma \in [1,2]$ for $\epsilon$-graph and full-RBF graphs. For the similarity network, we can then make the following remarks for the asymptotic scenario:

\begin{enumerate}
  \item While cut-ratio $q$ varies with graph construction, the imbalance coefficient $y$ is invariant. In particular we expect $q$ for $k$-NN to be larger relative to $q$ for full-RBF and $\epsilon$-graphs since $\gamma < 1$.

  \item We expect PCut to output similar results for all graph constructions. This follows from the limit-cut behavior and the limiting independence of $y$ to graph construction.

  \item We can loosely say that if data is imbalanced and with sufficiently proximal clusters then asymptotically $k$-NN, full-RBF and $\epsilon$-graphs can all fail when RCut is minimized. To see this consider an imbalanced mixture of two Gaussians. 
By suitably choosing the means and variances we can construct sufficiently proximal clusters with same imbalance but relatively large $q$ values. This is because $f(x)$ will be relatively large even at density valleys for proximal clusters. Our statement then follows from Remark \ref{remark}.
\end{enumerate}

Similar to the similarity network, we can build intuition for the performance of SC in the case of imbalanced clusters using the stochastic block model (SBM) as the generative process for a connectivity network. The stochastic block model is widely used for community detection to parameterize the problem using edge existence probabilities within and between communities. Consider a network with two communities of sizes $n_1$, $n_2$ where for this case we have $\alpha = \frac{\min\{n_1, n_2\}}{n} = \frac{n_1}{n}$ w.l.o.g.~and assume subnetworks for the communities are generated by the Erd\H{o}s-R\'enyi graph with internal edge existence probabilities $p_1$, $p_2$ and edge probability $q$ between two nodes in different subnetworks. 

Using the phase transition analysis proposed by \cite{hero}, we observe that an asymptotic lower bound $q_{LB}$ to $q$ (i.e.~if $q < q_{LB}$ SC recovers clusters almost surely) is $q_{LB} = \frac{\alpha p_1 + (1-\alpha)p_2 - |\alpha p_1 - (1-\alpha) p_2|}{2 (1-\alpha)}$ while an asymptotic upper bound $q_{UB}$ (i.e.~if $q > q_{UB}$ SC fails to recover clusters almost surely) is given by $q_{UB} = \frac{\alpha p_1 + (1-\alpha)p_2 - |\alpha p_1 - (1-\alpha) p_2|}{2 \alpha}$. The gap between the two bounds widens as $\alpha$ approaches zero and the lower bound shrinks linearly with $\alpha$ for most scaling regimes of interest, e.g.~when $\alpha p_1 = (1-\alpha)p_2$. This result implies that the performance of spectral clustering based community detection is more uncertain and tends to be worse when trying to detect small communities.

In summary, for similarity networks we have learned that the optimal RCut/NCut depend on graph construction and can fail for imbalanced proximal clusters for $k$-NN, $\epsilon$-graph, full-RBF constructions on same data. PCut is computationally intractable but asymptotically invariant to graph construction and picks the right solution. Since SC is a relaxed variant of optimal RCut/NCut, we can expect it to have similar behavior to the optimal RCut/NCut.
Similarly, we argued that the SC performance on a given graph is expected to deteriorate for small clusters in connectivity networks. This motivates the following section.

\subsection{Using spectral clustering for PCut} 
\label{subsec:sol}

For the data clustering/SSL problem, the discussion in Sec.~\ref{subsec:alg} suggests the possibility of controlling cut-ratio $q$ through modification of the underlying graph parameters while not impacting $y$ (which is invariant to different graph constructions). This key insight leads us to the following framework for solving PCut:

\begin{itemize}
\item [(A)]
{\bf Parameter optimization:}  Generate several candidate optimal RCuts/NCuts as a function of graph parameters. Pose PCut over these candidate cuts rather than arbitrary cuts as in Eq.~\eqref{e.empopt}. Thus PCut is now parameterized over graph parameters.
\item [(B)]
{\bf Graph parameterization:} If the graph parameters are not sufficiently rich to allow for adaptation to imbalanced or proximal cuts, (A) would be useless. Therefore, we want graph parameterizations that allow sufficient flexibility such that the posed optimization problem is successful for a broad range of imbalanced and proximal data.
\end{itemize}

We first consider the second objective. For the similarity network, we have found in our experiments (cf.~Sec.~\ref{sec:experiment}) that the parameterization based on RBF $k$-NN graphs is not sufficiently rich to account for varying levels of imbalanced and proximal data. To induce even more flexibility we introduce a new parameterization that we also generalize to connectivity networks.

\vspace{5pt}
\noindent
{\bf Rank modulated degree (RMD) graphs:} \\
We introduce RMD graphs that are a richer parameterization of graphs that allow for more control over $q$ and offers sufficient flexibility in dealing with a wide range of imbalanced and proximal data.
Our framework adaptively modulates the node degrees on the given baseline graph, while selectively removing edges in low density regions and adding in high-density regions. This modulation scheme is based on a novel ranking scheme for data samples introduced in Sec.~\ref{sec:RMD_idea}, which reflect the relative density and allows the identification of high/low density nodes. 
For similarity networks, we consider $k$-NN graphs since it is easier to ensure graph connectivity compared to $\epsilon$-graphs. 

For connectivity networks, we adopt a similar scheme that selectively removes edges from the given graph to improve SC performance. We again aim to remove edges between the clusters (``low density'' regions), while keeping edges that are present inside the clusters (``high density'' regions). Since we do not have similarity measures between nodes, we use other metrics to choose how many and which edges to remove for a node. 

We are now left to pose PCut over graph parameters or candidate cuts, which we describe in detail in the following section. For similarity networks we construct a universal baseline graph for the purpose of comparison among different cuts and to pick the cut that solves Eq.~\eqref{e.empopt}. These different cuts are obtained by means of SC and are parameterized by graph construction parameters. PCut is then solved on the baseline graph over candidate cuts realized from SC.
The optimization framework is illustrated in Fig.~\ref{f.framework} in the context of data clustering.

\section{Our Algorithm}\label{sec:RMD_idea}

\subsection{Similarity networks}

Given $n$ data samples, our task is unsupervised clustering or SSL, assuming the number of clusters/classes $K$ is known. We start with a baseline $k_0$-NN graph $G_0=(V,E_0)$ built on these samples with $k_0$ large enough to ensure graph connectivity.
Main steps of our PCut framework are as follows.

\begin{table}[h]
  \centering
  \begin{tabular}{l}
    \hline
    \textbf{ Main Algorithm:\, RMD Graph-based PCut } \\
    \hline
    1. Compute the rank $R(x_v)$ of each sample $x_v,i=1,...,n$; \\
    2. For different configurations of parameters, \\
      \,\,\, a. Construct the parametric RMD graph; \\
      \,\,\, b. Apply spectral methods to obtain a $K$-partition on the current RMD graph;  \\
    3. Among various partition results, pick the ``best" (evaluated on baseline $G_0$). \\
    \hline
  \end{tabular}
  \label{tab:main_alg}
\end{table}

\begin{figure}[htbp]
  \centering
  \includegraphics[width=0.5\textwidth]{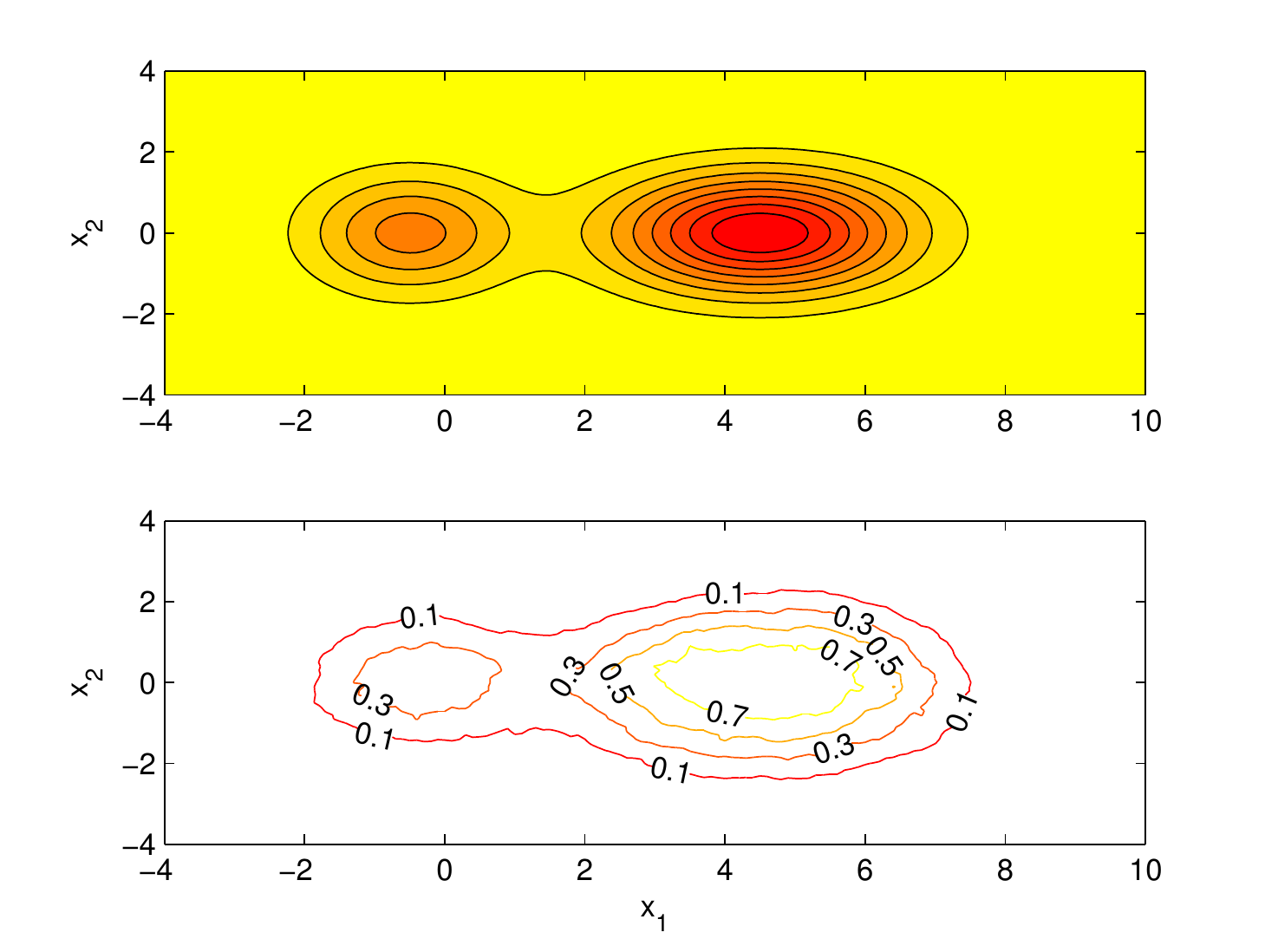}
  \caption{\small Density level sets \& rank estimates computed from samples for a mixture of two Gaussians.}
  \label{f.pdf_rank}
\end{figure}

\vspace{5pt}
\noindent
{\bf (1) Rank computation:} \\
We compute the rank $R(v) = R(x_v)$ of every node $v$ as follows:
\begin{equation}\label{eq:grank}
  R(x_v) = \frac{1}{n}\sum_{w \in V} \mathbb{I}_{\{\eta(x_v)\leq \eta(x_w)\}},
\end{equation}
where $\mathbb{I}$ denotes the indicator function and $\eta(x_v)$ is a statistic reflecting the relative density at node $v$. Since $f$ is unknown, we choose average nearest neighbor distance as a surrogate for $\eta$. To this end let $N(v)$ be the set of all neighbors for node $v \in V$ on the baseline graph, and we let 
\begin{equation}\label{eq:grank1}
  \eta(x_v)=\frac{1}{|N(v)|}\sum_{w \in N(v)} \|x_v - x_w\|.
\end{equation}
The ranks $R(x_v) \in [0, 1]$ are relative orderings of samples and are uniformly distributed.
$R(x_v)$ indicates whether a node $v$ lies near density valleys or high-density areas, as illustrated in Fig.~\ref{f.pdf_rank}.

\vspace{5pt}
\noindent
{\bf (2a) Parameterized graph construction:} \\
We consider three parameters, $\lambda \in [0,\,1]$, $k$ for $k$-NN and $\sigma$ for RBF similarity. These are then suitably discretized. We generate a weighted graph $G(\lambda,k,\sigma)=(V, E(\lambda,k,\sigma),W(\lambda,k,\sigma))$ on the same node set as the baseline graph but with different edge sets. For each node $v \in V$ we construct edges to its $k_\lambda(v)$ nearest neighbors, with $k_\lambda(v)$ given by
\begin{equation}\label{eq:degree}
  k_{\lambda}(v) = k(\lambda+2(1-\lambda)R(x_v)),
\end{equation}
which results in the RMD parameterization through different $\lambda$. Note that $\lambda = 1$ corresponds to no degree modulation. For non-RMD parameterizations that we compare to (such as RBF $k$-NN) we vary $k$ and $\sigma$.

\vspace{5pt}
\noindent
{\bf (2b) Obtaining cuts on the parameterized graphs:} \\
From $G(\lambda,k,\sigma)$ we generate a family of $K$ partitions $C_1(\lambda,k,\sigma), C_2(\lambda,k,\sigma), \ldots, C_K(\lambda,k,\sigma)$. These cuts are generated based on the eventual learning objective. For instance, if $K$-clustering is the eventual goal these $K$-cuts are generated using SC. For SSL we use RCut-based Gaussian random fields (GRF) and NCut-based graph transduction via alternating minimization (GTAM) to generate cuts. These algorithms all involve minimizing RCut/NCut as the main objective (SC) or some smoothness regularizer (GRF, GTAM). For details about these algorithms readers are referred to \cite{Zhu08,WanJebCha08,Luxburg07,Chung96}.

\vspace{5pt}
\noindent
{\bf (3) Parameter optimization:} \\
The final step is to solve Eq.~\eqref{e.empopt} on the baseline graph $G_0$.
We assume prior knowledge that the smallest cluster is at least of size $\delta n$.
The $K$-partitions obtained from step (2b) are parameterized as $\left(C_1(\lambda,k,\sigma),...,C_K(\lambda,k,\sigma)\right)$.
We optimize over the parameters $(\lambda,k,\sigma)$ to obtain the minimum cut partition on $G_0$
\begin{equation}\label{eq:selection}
  \min_{\lambda,k,\sigma} \{Cut_0\left(C_1,...,C_K\right) = \sum^K_{i=1} Cut_0(C_i,\bar{C}_i)\}~~~\mathrm{s.t.}~~~\min\{|C_1(\lambda,k,\sigma)|,...,|C_K(\lambda,k,\sigma)|\}\geq \delta n,
\end{equation}
where $Cut_0(\cdot)$ denotes evaluating cut values on the baseline graph $G_0$. Partitions with clusters smaller than $\delta n$ are discarded.

\vspace{5pt}
\noindent
{\bf Remarks:}
\begin{enumerate}
  \item While step (3) suggests a grid search over several parameters, it turns out that other parameters such as $k,\sigma$ do not play an important role as $\lambda$. Indeed, the experiments in Sec.~\ref{sec:experiment} show that while step (3) can select appropriate $k,\sigma$, it is by searching over $\lambda$ that adapts spectral methods to data with varying levels of imbalancedness (cf.\ Table \ref{tab:real_SC}, RBF k-NN vs.\ RBF RMD).
  \item Our framework uses existing spectral algorithms and thus can be combined with other graph-based partitioning algorithms to improve performance for imbalanced data, such as 1-spectral clustering, sparsest cut or minimizing conductance \cite{Buhler09,Hein10,Szlam10,Arora09}. We utilize SC for data clustering and GRF/GTAM algorithms for the SSL problem in our experiments in Sec.~\ref{sec:experiment}, with the same RMD graph parameterization framework.
\end{enumerate}

\subsection{Connectivity networks}
\label{subsec:rmd_c}

We adapt the rank computation framework and the degree modulation scheme from similarity networks to the connectivity networks case. Since we do not have access to similarity scores between nodes such as distances as in similarity networks, it is not possible to directly adapt the computation of the score function $\eta(v)$ using similarities between nodes. To this end, we adopt the count of common neighbors metric as a similarity indicator between two nodes. This statistic is defined by $s(v,w) = |{\cal N}(v) \cap {\cal N}(w)|$ where ${\cal N}(v)$ denotes the set of neighbors of a node $v$ and is used frequently as a heuristic measure of similarity in applications such as link prediction \cite{link}. One interesting application of the statistic is community detection without spectral clustering, where \cite{common} considers the scenario with exactly balanced community sizes and aims to discover clusters directly using the statistic. In contrast, we focus on clusters with imbalanced sizes and use the statistic only as a similarity measure to construct an analogy to the similarity network case.
The intuition about the count of common neighbors statistic is that two nodes that are in the same cluster share more neighbors (which are mostly from the same cluster) than two nodes from different clusters. Thus it can be used as a measure to determine whether or not two nodes belong to the same cluster. 

To extend the RMD framework to the connectivity network, we again compute the rank of a node $R(v)$ from a relative ``density'' function $\eta(v)$, which for this case we define as
\[ \eta(v) = -\frac{1}{|{\cal N}(v)|} \sum_{w \in {\cal N}(v)} s(v,w), \]
where we essentially replaced the Euclidean distance $\|x_v - x_w\|$ with the negative similarity $-s(v,w)$. Then using this relative density measure, the rank $R(v)$ is computed as in Eq.~\eqref{eq:grank1}. In this context, the nodes with high rank are ``high density'' nodes, where they are connected more frequently to their own clusters than to different clusters compared to the average node in the graph. On the other hand ``low density'' nodes with low rank are connected more to other clusters compared to the average case.

Given the rank of a node, we modulate the node's degree using the formula
\[ d_\lambda(v) = d(v) (\lambda + (1-\lambda)R(v)), \]
where we differ from Eq.~\eqref{eq:degree} by not multiplying the $(1-\lambda) R(v)$ term by 2. This is because we only \emph{decrease} the degrees of the nodes by removing edges, rather than increasing or decreasing according to rank. One remaining issue is which edges to remove from a node given that its new degree $d_\lambda(v)$ is less than original degree $d(v)$. Considering the analogy to the $k$-NN graph, we remove the $d_\lambda(v)-d(v)$ edges that are connected to neighbors farthest from $v$, i.e.~for which the count of common neighbors is smallest. This procedure prioritizes the removal of edges to neighbors in other clusters before the edges that connect to neighbors in the same cluster. In addition, more edges are removed from nodes with lower rank, i.e.~nodes which connect to nodes in other clusters more frequently than the average. Similarly, less edges are removed from nodes with higher rank, i.e.~nodes that do not connect to other clusters as frequently. 

The parameterization and parameter optimization in parts (2b) and (3) follow as in the similarity network case, where the only search parameter we use is $\lambda$ and not other parameters such as $k$ or $\sigma$. We note that it would be possible to determine the new degree of a node in a more robust manner given parameters such as the cluster size imbalance $\alpha$ and the probabilities $p_1$, $p_2$ and $q$ in the stochastic block model considered in Sec.~\ref{subsec:alg}, however we do not assume knowledge of these parameters and instead use the rank of a node and the parameterization over $\lambda$ to account for their uncertainty.

\section{Analysis of RMD for Similarity Networks}
\label{sec:thm}

We now present an asymptotic analysis for binary cuts in similarity networks that shows how RMD helps control the cut-ratio $q$, introduced in Sec.~\ref{sec:motiv}. We remark that since we analyze the limit cut behavior of RCut/NCut that is directly related to SC, it may be possible to extend it to other methods such as GTAM for SSL that are based on the NCut objective.

Assume the dataset $\{x_1,\ldots,x_n\}$ is drawn i.i.d.\ from an underlying density $f$ in $\mathbb{R}^d$. Let $G=(V,E)$ be the unweighted RMD graph. Given a separating hyperplane $S$, denote with $C^+, C^-$ the two subsets of $C$ split by $S$ and let $\eta_d$ denote the volume of the unit ball in $\mathbb{R}^d$. Assume the density $f$ satisfies the regularity conditions stated below.

\emph{Regularity conditions:} $f(\cdot)$ has a compact support, and is continuous and bounded: $f_{max} \geq f(x) \geq f_{min}>0$. It is smooth, i.e.\ $||\nabla f(x)||\leq\lambda$, where $\nabla f(x)$ is the gradient of $f(\cdot)$ at $x$. There are no flat density regions, i.e.\ $\mathcal{P}\left\{y: |f(y)-f(x)|<\sigma\right\}\leq M\sigma$ for all $x$ in the support and $\sigma>0$, where $M$ is a constant.

First we show the asymptotic consistency of the rank $R(y)$ at some point $y$. The limit of $R(y)$ is $p(y)$, which is defined as the complement of the volume of the level set containing $y$. Note that $p$ exactly follows the shape of $f$ and is always between $[0,1]$ no matter how $f$ scales.
\begin{theorem}\label{rank-pvalue}
Assume $f(x)$ satisfies the above regularity conditions. As $n\rightarrow\infty$, we have
\begin{equation}
  R(y) \longrightarrow p(y):= \int_{ \left\{ x:f(x) \leq f(y) \right\}} f(x) \dx x.
\end{equation}
\end{theorem}

This theorem implies that the rank $R(y)$ of a point $y$ is a good estimate of $p(y)$, which is in turn related to the shape of the density $f(y)$. Thus $R(y)$ is a useful metric for identifying high and low density points which is necessary for modulating node degrees to emphasize density valleys in the RMD framework.

The proof involves the following two steps:
\begin{itemize}
  \item[1.] The expectation of the empirical rank $\mathbb{E}\left[R(y)\right]$ is shown to converge to $p(y)$ as $n\rightarrow\infty$.
  \item[2.] The empirical rank $R(y)$ is shown to concentrate at its expectation as $n\rightarrow\infty$.
\end{itemize}
Details can be found in the appendix.
Small/large $R(x)$ values correspond to low/high density respectively. $R(x)$ asymptotically converges to an integral expression, so it is smooth (Fig.~\ref{f.pdf_rank}). Also $p(x)$ is uniformly distributed in $[0,1]$, which makes it appropriate to modulate the degrees with control of minimum, maximum and average degrees.

Next we study RCut and NCut values induced on the unweighted RMD graph. We assume for simplicity that each node $v$ is connected to exactly $k_\lambda(v)$ nearest neighbors given by Eq.~\eqref{eq:degree}. The limit cut expression on RMD graph involves an additional adjustable term which varies point-wise according to the density.

\begin{theorem}\label{part2}
Assume $f$ satisfies the above regularity conditions and also the general assumptions in \cite{Maier1}. Let $S$ be a fixed hyperplane in $\mathbb{R}^d$. For an unweighted RMD graph, set the degrees of points according to Eq.~\eqref{eq:degree}, where $\lambda \in (0,1)$ is a constant. Let $\rho(x) = \lambda + 2(1-\lambda)p(x)$. Assume $k_n/n \rightarrow 0$. Assume $k_n/\sqrt{n}\rightarrow\infty$ if $d=1$ and assume $k_n/\log{n}\rightarrow\infty$ if $d \geq 2$. Then as $n\rightarrow\infty$ we have that
\begin{equation} \label{eq:rcut}
    \frac{1}{k_n}\sqrt[d]{\frac{n}{k_n}}RCut_n(S) \longrightarrow  C_d B_S \int_S{f^{1-\frac{1}{d}}(s)\rho^{1+\frac{1}{d}}(s) \dx s},
\end{equation}
\begin{equation} \label{eq:ncut}
    \sqrt[d]{\frac{n}{k_n}}NCut_n(S) \longrightarrow  C_d B_S \int_S{f^{1-\frac{1}{d}}(s)\rho^{1+\frac{1}{d}}(s) \dx s},
\end{equation}
where $C_d = \frac{2\eta_{d-1}}{(d+1) \eta_d^{1+\frac{1}{d}}}$, $B_S = \frac{1}{\mu(C^+)} + \frac{1}{\mu(C^-)}$ and $\mu(C^{\pm})=\int_{C^{\pm}}f(x) \dx x$.
\end{theorem}

The proof shows the convergence of the cut term and balancing term respectively:
\begin{align}
  \frac{1}{nk_n}\sqrt[d]{\frac{n}{k_n}}cut_n(S)
  \longrightarrow C_d\int_S{f^{1-\frac{1}{d}}(s)\rho^{1+\frac{1}{d}}(s) \dx s}, \\
  n\frac{1}{|V^\pm|}\longrightarrow
  \frac{1}{\mu(C^\pm)}, \,\,\,\,  nk_n\frac{1}{vol(V^\pm)}\longrightarrow
  \frac{1}{\mu(C^\pm)}. \label{eq:term2N1}
\end{align}
The analysis is an extension of \cite{Maier1} and the proof is provided in the appendix.

Theorem \ref{part2} shows that the RMD parameterization can affect the RCut/NCut behavior in a meaningful manner as the densities are modulated with the $\rho^{1+\frac{1}{d}}(s)$ term that varies with parameter $\lambda$. We discuss the effects of this modulation on imbalanced data next.

\vspace{5pt}
\noindent
{\bf Imbalanced data \& RMD graphs:} \\
In the limit cut behavior, without our $\rho$ term, the balancing term $B_S = \frac{1}{\alpha(1-\alpha)}$ could induce a larger RCut/NCut value for the density valley cut than the balanced cut when the underlying data is imbalanced, i.e.\ $\alpha$ is small.
Applying our parameterization scheme inserts an additional term $\rho(s)=(\lambda+2(1-\lambda)p(s))$ in the limit-cut expressions.
$\rho(s)$ is monotonic in the $p$-value and thus allows the cut-value to be further reduced/increased at low/high density regions. Indeed for small $\lambda$ values, cuts $S$ near peak densities have $p(s)\approx 1$ and so $\rho^{1+\frac{1}{d}}(s) \approx 2^{1+\frac{1}{d}}$, while near valleys we have $\rho^{1+\frac{1}{d}}(s) \approx \lambda^{1+\frac{1}{d}} \ll 1$. This has a direct bearing on the cut-ratio $q$ since small $\lambda$ can reduce the cut-ratio $q$ for a given $y$ (see Fig.~\ref{fig:qy}) and leads to better control of cuts on imbalanced data. In summary, this analysis shows that RMD graphs used in conjunction with the optimization framework of Fig.~\ref{f.framework} can adapt to varying levels of imbalanced data.

\section{Experiments}
\label{sec:experiment}
Experiments in this section involve both synthetic and real datasets, where we consider data clustering and semi-supervised learning with similarity networks for the first two subsections and consider community detection with connectivity networks in the third subsection. For the similarity network problems, we focus on imbalanced data by randomly sampling from different classes disproportionately.  For comparison purposes we compare the RMD graph with full-RBF, $\epsilon$-graph, RBF $k$-NN, $b$-matching graph \cite{JebWanCha09} and full graph with adaptive RBF (full-aRBF) \cite{Zelnik04}. We view each as a parametric family of graphs parameterized by their relevant parameters and optimize over different parameters as described in Sec.~\ref{sec:RMD_idea} and Eq.~\eqref{eq:selection}. For RMD graphs we also optimize over $\lambda$ in addition. Error rates are averaged over 20 trials.

For clustering experiments we apply both RCut and NCut, but focus mainly on NCut for brevity as NCut is generally known to perform better. We report performance by evaluating how well the cluster structures match the ground truth labels, as is the standard criterion for partitional clustering \cite{xu05}. For instance consider Table 1 where error rates for USPS symbols 1, 8, 3, 9 are tabulated. We follow our procedure outlined in Sec.~\ref{sec:RMD_idea} and find the optimal partition that minimizes Eq.~\eqref{eq:selection} \emph{agnostic} to the correspondence between samples and symbols. Errors are then reported by looking at mis-associations.

For SSL experiments we randomly pick labeled points among imbalanced sampled data, guaranteeing at least one labeled point from each class. SSL algorithms such as RCut-based GRF and NCut-based GTAM are applied on parameterized graphs built from partially labeled data and generate various partitions. Again we follow our procedure outlined in Sec.~\ref{sec:RMD_idea} and find the optimal partition that minimizes Eq.~\eqref{eq:selection} agnostic to ground truth labels. Then labels for unlabeled data are predicted based on the selected partition and compared against the unknown true labels to produce the error rates.

\vspace{5pt}
\noindent {\it Time complexity:}
RMD graph construction has time complexity $O(d n^2 \log n)$ (similar to the $k$-NN graph). Computing cut value and checking cluster size for a partition takes $O(n^2)$. So if $D$ graphs are parameterized in total and the complexity of the learning algorithm is $T$, the overall time complexity is $O(D(d n^2 \log n + T))$.

\vspace{5pt}
\noindent {\it Tuning parameters:} Note that parameters including $\lambda,k,\sigma$ that characterize the graphs are variables to be optimized in Eq.~\eqref{eq:selection}. The remaining parameters are (a) $k_0$ in the baseline graph which is fixed to be $\sqrt{n}$, (b) imbalanced size threshold $\delta$ which is a priori fixed to be about $0.05$, i.e., 5\% of all samples.

\vspace{5pt}
\noindent {\it Evaluation against oracle:} To evaluate the effectiveness of our framework and the RMD parameterization, we compare against an oracle that is tuned to both ground truth labels as well as imbalance proportions.

\subsection{Synthetic illustrative data clustering example}

\begin{figure*}[tb]
  \centering
  \begin{minipage}[t]{.24\textwidth}
  \includegraphics[width = 1\textwidth]{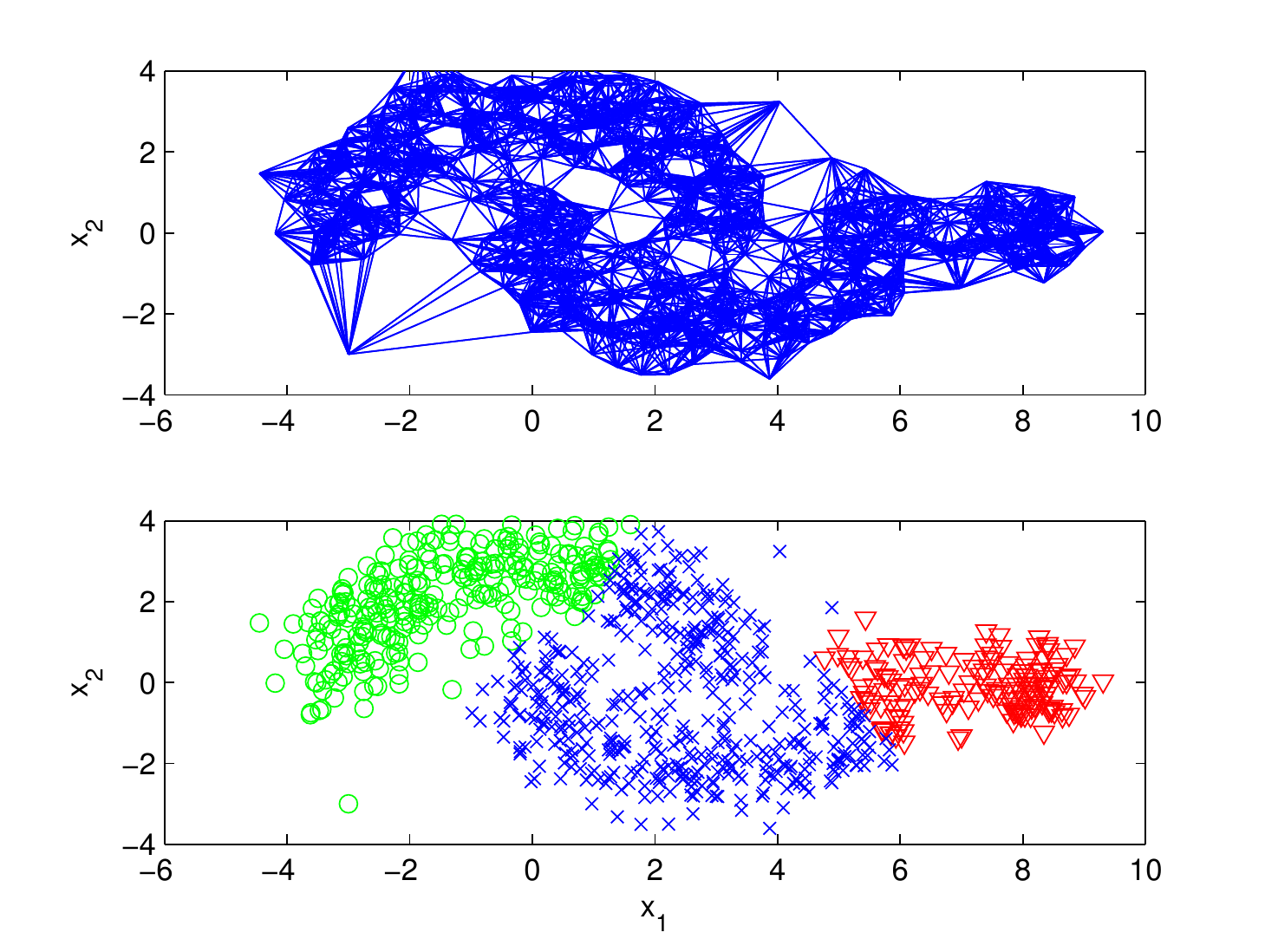}
  \makebox[4cm]{(a) $k$-NN}
  \end{minipage}
  \begin{minipage}[t]{.24\textwidth}
  \includegraphics[width = 1\textwidth]{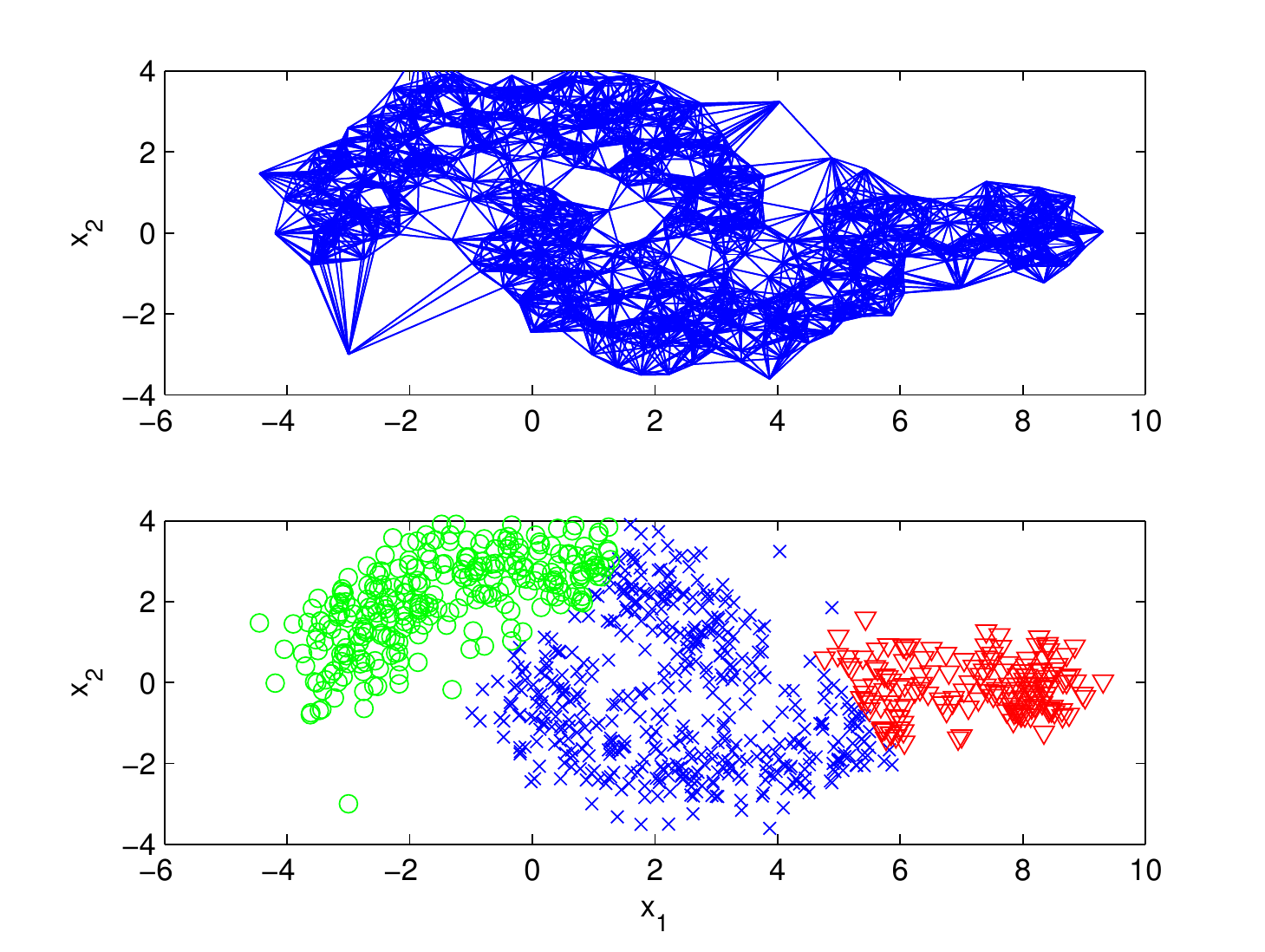}
  \makebox[4cm]{(b) $b$-matching}
  \end{minipage}
  \begin{minipage}[t]{.24\textwidth}
  \includegraphics[width = 1\textwidth]{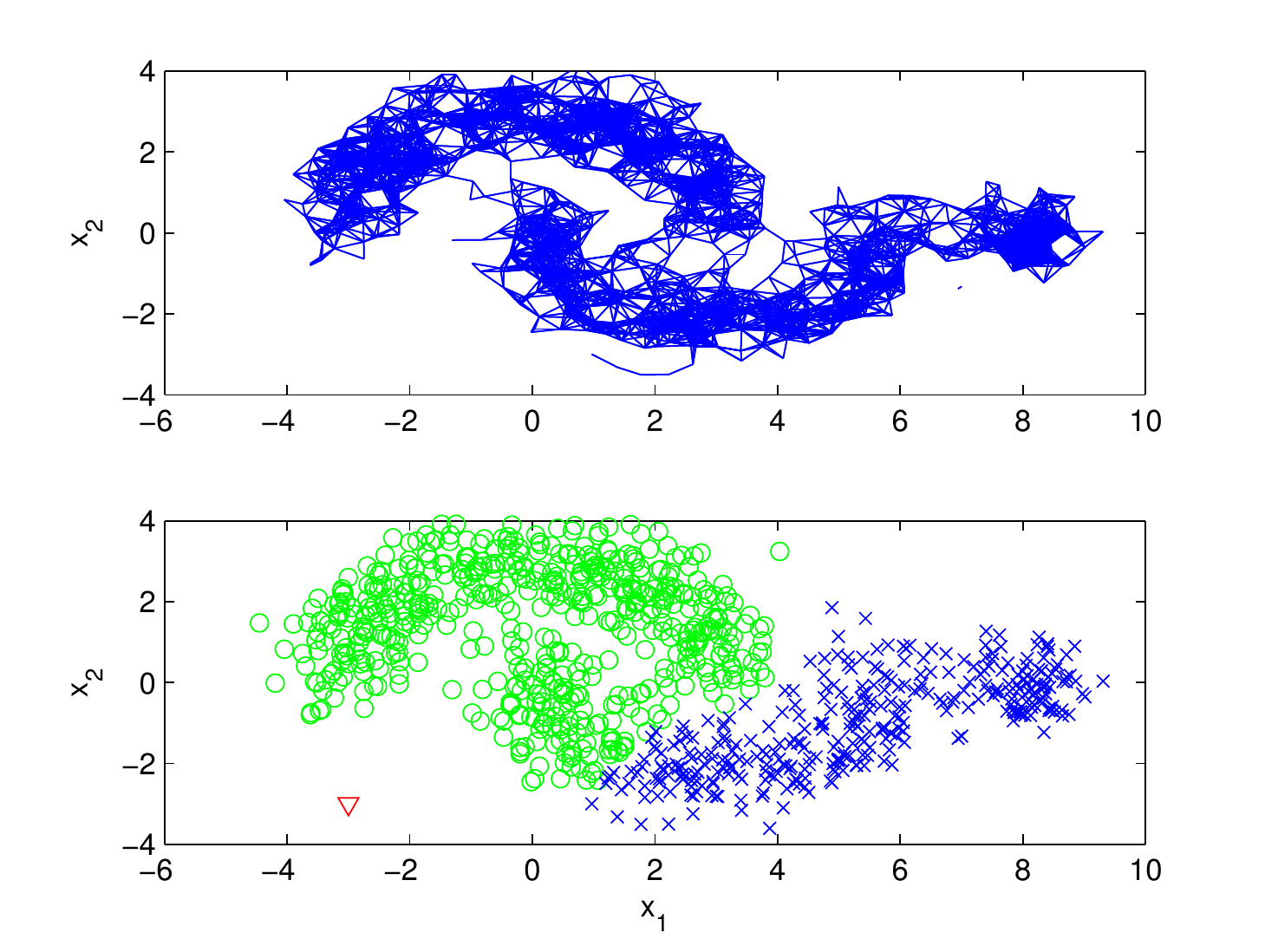}
  \makebox[5cm]{(c) $\epsilon$-graph (full-RBF)}
  \end{minipage}
  \begin{minipage}[t]{.24\textwidth}
  \includegraphics[width = 1\textwidth]{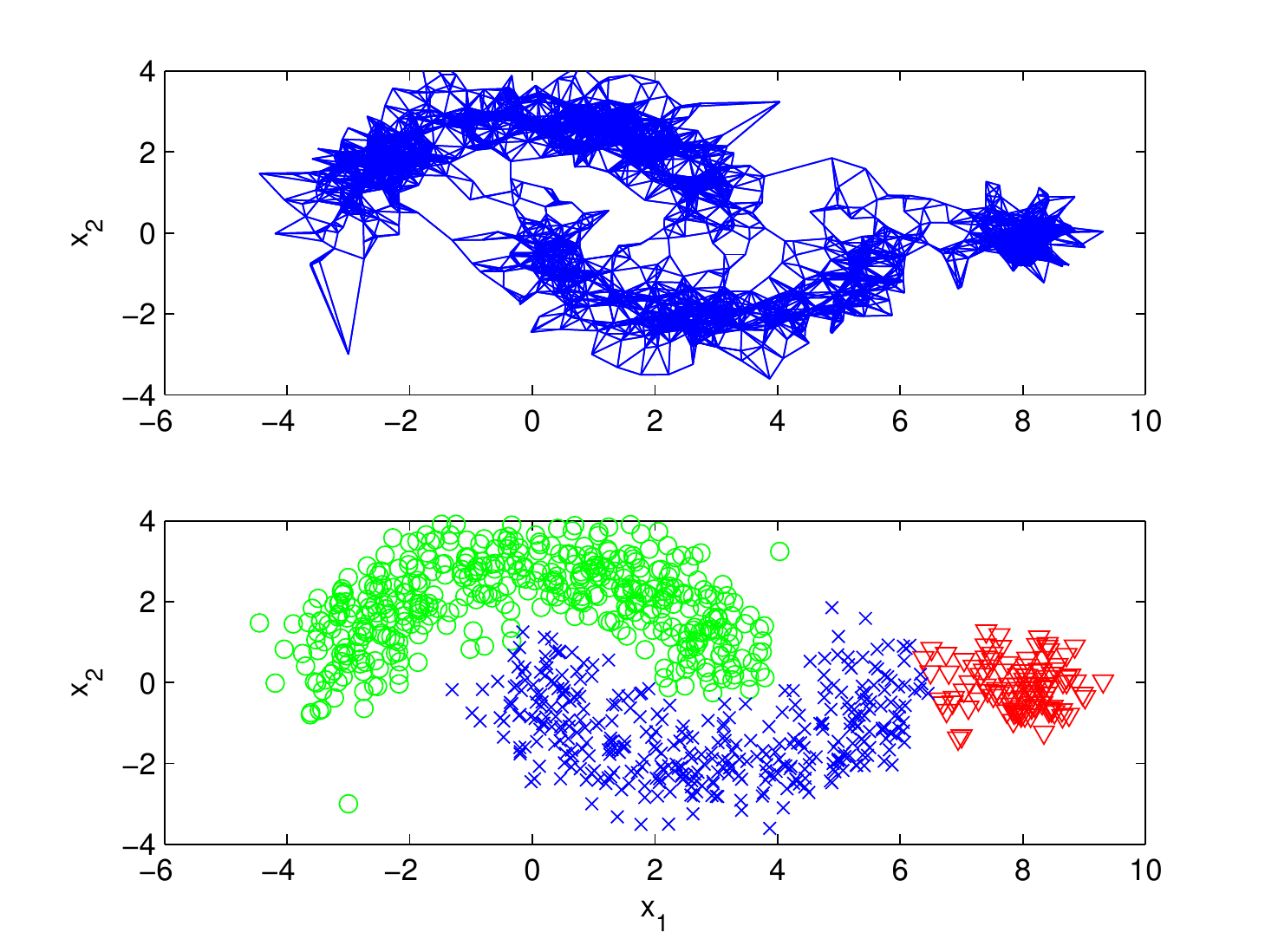}
  \makebox[4cm]{(d) RMD}
  \end{minipage}
  \caption{Clustering results of 3-partition SC on the two crescents and one Gaussian dataset. SC on $\epsilon$-graph completely fails due to the outlier. For $k$-NN and $b$-matching graphs SC cannot recognize the long winding low-density regions between the two crescents and fails to find the rightmost small cluster. Our method sparsifies the graph at low-density regions, allowing to cut along the valley, detect the small cluster and is robust to outliers.}
  \label{fig:complex_shape}
\end{figure*}

We consider a multi-cluster complex-shaped dataset which is composed of 1 small Gaussian and 2 crescent-shaped proximal clusters that is shown in Fig.~\ref{fig:complex_shape}. We have a sample size of $n=1000$ with the rightmost small cluster formed by $10\%$ of the samples and two crescents $45\%$ each. This example is only for illustrative purpose with a single run, so we did not parameterize the graph or apply the optimization step (3) in the framework. We fix $\lambda = 0.5$ and choose $k=l=30$, $\epsilon=\sigma=\tilde{d}_k$, where $\tilde{d}_k$ is the average $k$-NN distance. Model-based approaches can fail on such a dataset due to the complex shapes of clusters. The 3-partition SC based on RCut is applied. We observe in Fig.~\ref{fig:complex_shape} that on $k$-NN and $b$-matching graphs SC fails for two reasons: (1) SC cuts at balanced positions and cannot detect the rightmost small cluster, (2) SC cannot recognize the long winding low-density regions between the two crescents because there are too many spurious edges and the cut value along the curve is large. SC fails on the $\epsilon$-graph (similar to full-RBF) because the outlier point forms a singleton cluster and also cannot recognize the low-density curve. Our RMD graph significantly sparsifies the graph at low-densities, enabling SC to cut along the valley, detect small clusters and reject outliers.

\subsection{Real experiments with similarity networks}

We focus on imbalanced settings for several real datasets. We construct $k$-NN, $b$-matching, full-RBF and RMD graphs all combined with RBF weights, but do not include the $\epsilon$-graph because of its overall poor performance \cite{JebWanCha09}.
Our sample size varies from 750 to 1500.
We discretize not only $\lambda$ but also $k$ and $\sigma$ to parameterize graphs.
We vary $k$ in $\{5,10,20,30,\ldots,100,120,150\}$.
While small $k$ may lead to disconnected graphs this is not an issue for us since singleton cluster candidates are ruled infeasible in PCut. Also notice that for $\lambda=1$, RMD graph is identical to the $k$-NN graph. For RBF parameter $\sigma$ it has been suggested to use a value on the same scale as the average $k$-NN distance $\tilde{d}_k$ \cite{WanJebCha08}. This suggests a discretization of $\sigma = 2^j \tilde{d}_k$ with $j=-3,\,-2,\ldots,\,3$. We discretize $\lambda \in [0,\,1]$ with steps of $0.2$.

In the model selection step Eq.~\eqref{eq:selection}, cut values of various partitions are evaluated on the same $k_0$-NN graph with $k_0=30$, $\sigma = \tilde{d}_{30}$ before selecting the min-cut partition. The true number of clusters/classes $K$ is assumed known. We assume meaningful clusters are at least $5\%$ of the total number of points, i.e.\ $\delta=0.05$. We set the GTAM parameter $\mu=0.05$ as in \cite{JebWanCha09} for the SSL tasks and each time 20 randomly labeled samples are chosen with at least one sample from each class.

\begin{figure*}[tb]
  \centering
  \begin{minipage}[t]{.32\textwidth}
    \includegraphics[width = 1\textwidth]{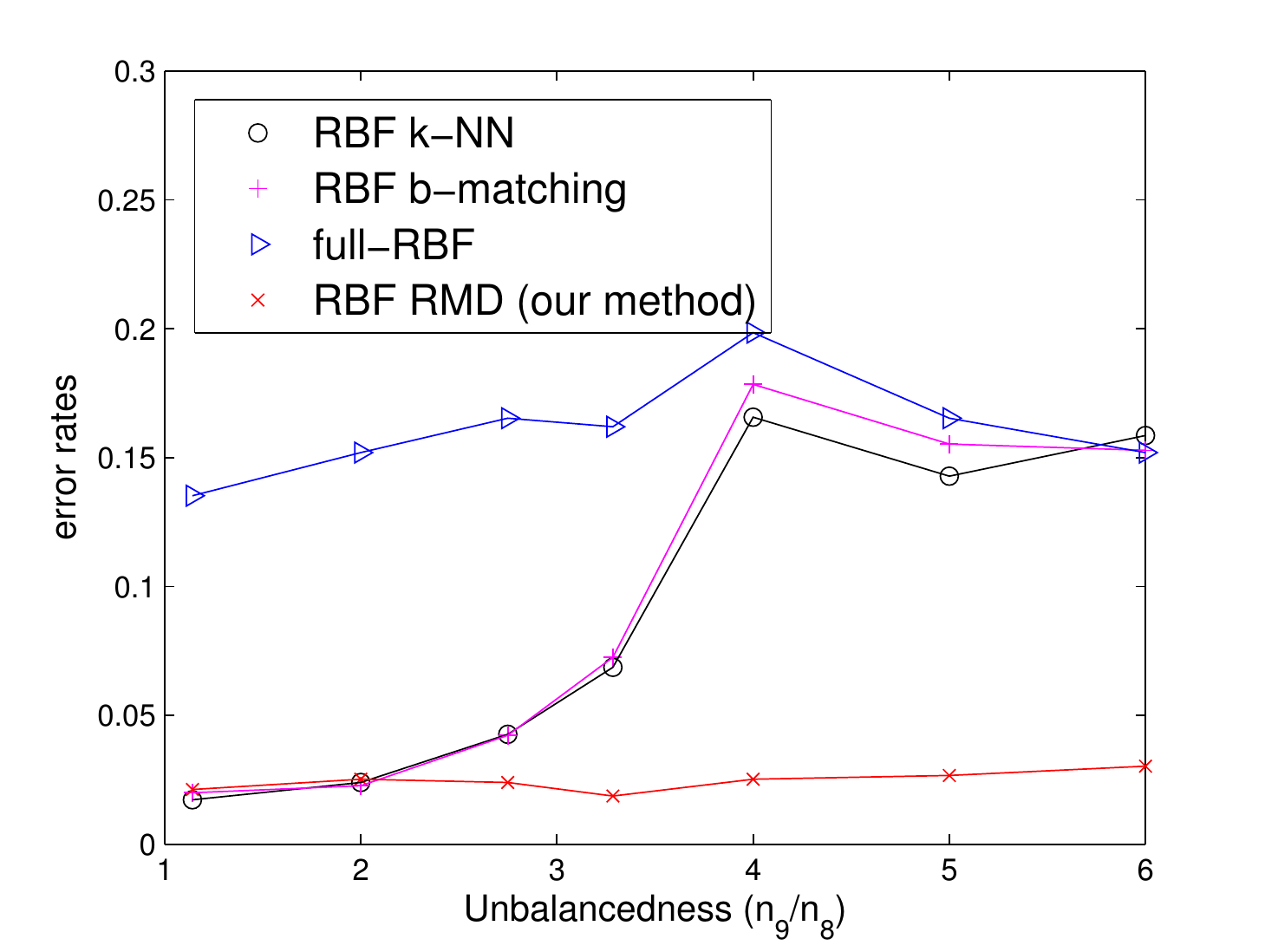}
    \makebox[5cm]{(a) SC (clustering)}
  \end{minipage}
  \begin{minipage}[t]{.32\textwidth}
    \includegraphics[width = 1\textwidth]{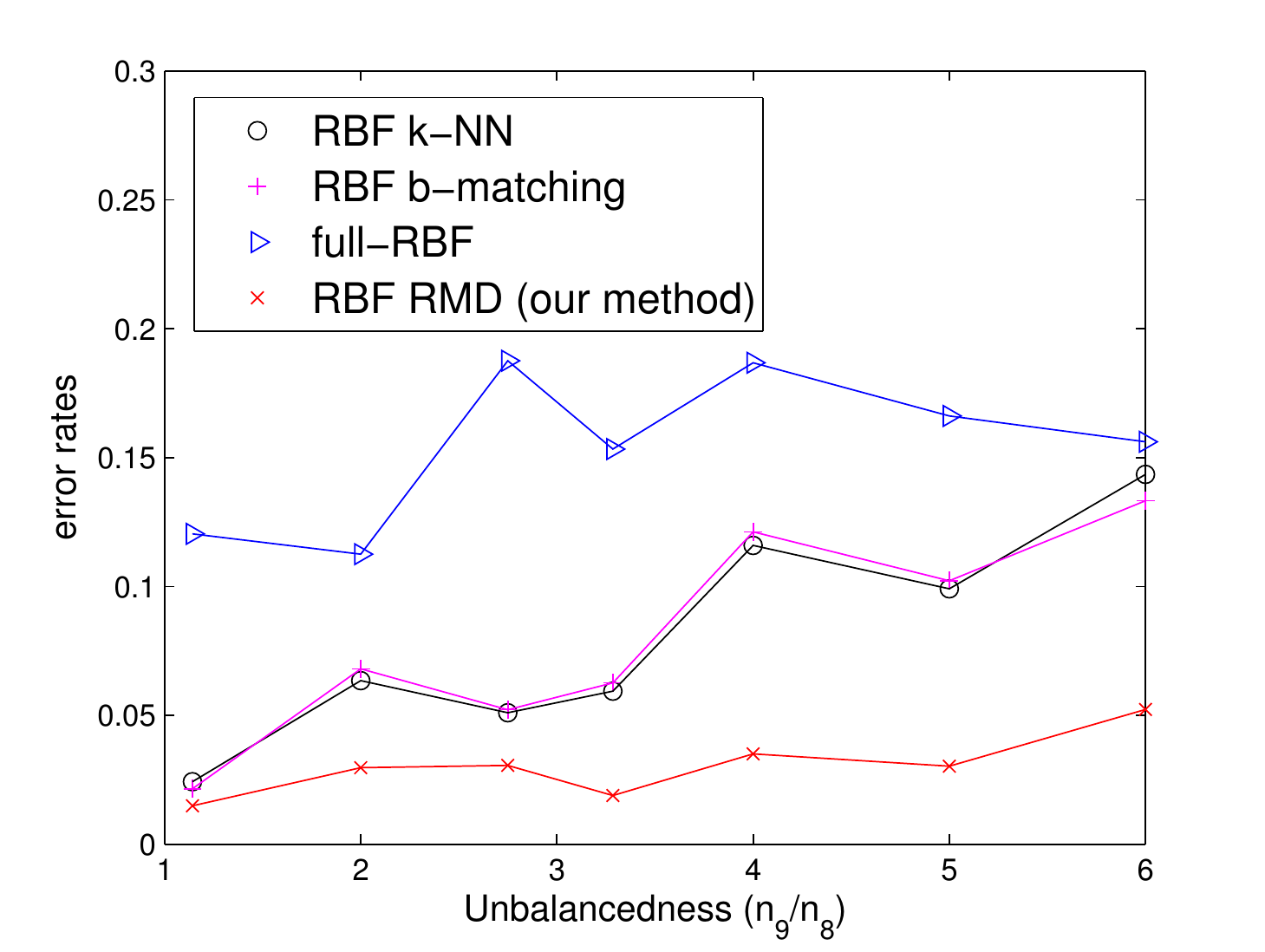}
    \makebox[5cm]{(b) GRF (SSL)}
  \end{minipage}
  \begin{minipage}[t]{.32\textwidth}
    \includegraphics[width = 1\textwidth]{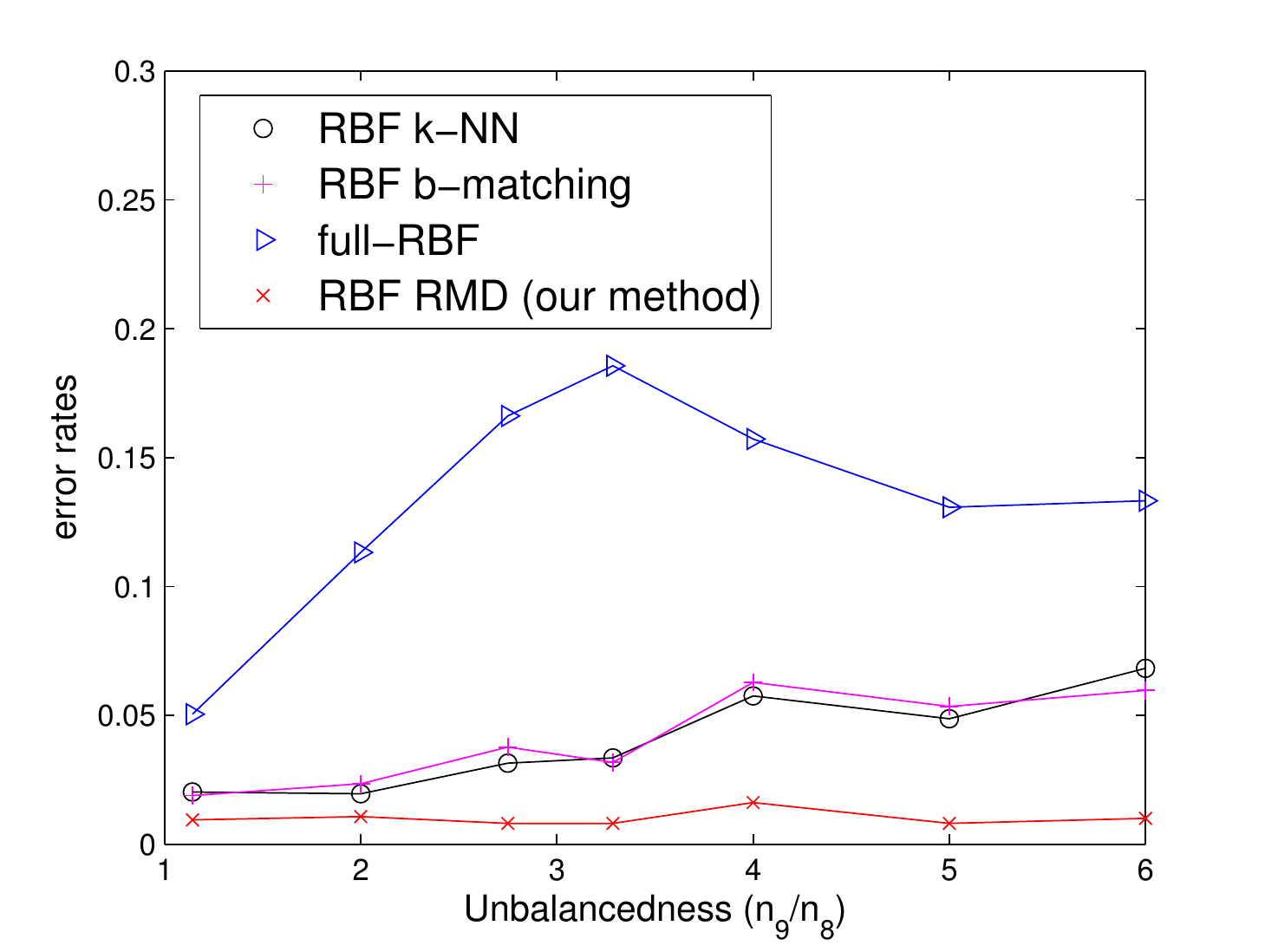}
    \makebox[5cm]{(c) GTAM (SSL) }
  \end{minipage}
  \caption{Error rates of SC and SSL algorithms on USPS 8vs9 with varying levels of imbalancedness. Our RMD scheme remains competitive when the data is balanced and adapts to imbalancedness much better than traditional graph constructions.}
  \label{fig:USPS8v9}
\end{figure*}

\vspace{5pt}
\noindent
\textbf{Varying imbalancedness:} \\
We use the digits 8 and 9 in the 256-dim USPS digit dataset and randomly sample 750 points with different levels of imbalancedness. Normalized SC, GRF and GTAM are then applied. Fig.~\ref{fig:USPS8v9} shows that when the underlying clusters/classes are balanced, our RMD method performs as well as traditional graphs. As the imbalancedness increases the performance of other graphs degrades, while our method can adapt to different levels of imbalancedness.
\begin{table}[tb]
\caption{ Imbalancedness of datasets. }
\centering
\begin{tabular}{|c|c|}
  \hline
  Datasets     & \#samples per cluster  \\
  \hline\hline
  2-cluster (e.g.\ USPS 8/9)     & 150/600 \\    \hline
  3-cluster (e.g.\ SatImg 3/4/5)  & 200/400/600    \\ \hline
  4-cluster (e.g.\ USPS 1/8/3/9)  & 200/300/400/500     \\
  \hline
\end{tabular}
\label{tab:Imbalanced_demo}
\end{table}

\begin{table*}[tb]
\caption{ Error rates of normalized SC on various graphs for imbalanced real datasets. Our method performs significantly better than other methods. First row (``BO'' Balanced Oracle) shows RBF $k$-NN results on imbalanced data with $k,\sigma$ tuned using ground truth labels but on balanced data. Last row (``O'' Oracle) shows the best oracle results of RBF RMD on imbalanced data. }
\vspace{-10pt}
\centering
\begin{tabular}{|c||c|c|c|c|c|c|c|c|c|c|}
  \hline
  \multirow{2}{*}{Error Rates (\%)}   &   \multicolumn{2}{c|}{USPS}  &   \multicolumn{3}{c|}{SatImg}  &   \multicolumn{3}{c|}{OptDigit}   & \multicolumn{2}{c|}{LetterRec} \\
  \cline{2-11}
  & 8vs9 & 1,8,3,9 & 4vs3 & 3,4,5 & 1,4,7 & 9vs8 & 6vs8 & 1,4,8,9 & 6vs7 & 6,7,8 \\
  \hline\hline
  RBF $k$-NN (BO)    & 33.20 & 17.60 & 15.76 & 22.08 & 25.28 & 15.17 & 11.15  & 30.02 & 7.85 & 38.70     \\
  RBF $k$-NN        & 16.67 & 13.21 & 12.80 & 18.94 & 25.33 & 9.67  & 10.76  & 26.76 & 4.89 & 37.72 \\
  RBF $b$-match  & 17.33 & 12.75 & 12.73 & 18.86 & 25.67 & 10.11  & 11.44  & 28.53 & 5.13 & 38.33 \\
  full-RBF          & 19.87 & 16.56 & 18.59 & 21.33 & 34.69 & 11.61 & 15.47 & 36.22 & 7.45 & 35.98 \\
  full-aRBF         & 18.35 & 16.26 & 16.79 & 20.15 & 35.91 & 10.88 & 13.27 & 33.86 & 7.58 & 35.27 \\
  RBF RMD           & 4.80  & 9.66 & 9.25 & 16.26 & 20.52 & 6.35  & 6.93  & 23.35 & 3.60 & 28.68 \\
  RBF RMD (O)        & 3.13  & 7.89 & 8.30 & 14.19 & 18.72 & 5.43  & 6.27  & 19.71 & 3.02 & 25.33 \\
  \hline
\end{tabular}
\label{tab:real_SC}
\end{table*}

\vspace{5pt}
\noindent
\textbf{Other real datasets:} \\
We apply SC and SSL algorithms on several other real datasets including USPS (256-dim.), Statlog landsat satellite images (4-dim.), letter recognition images (16-dim.) and optical recognition of handwritten digits (16-dim.) \cite{uci10}.
We sample the datasets in an imbalanced manner as shown in Table \ref{tab:Imbalanced_demo}.

In Table \ref{tab:real_SC}, the first row is the imbalanced results of RBF $k$-NN using oracle with $k, \sigma$ parameters tuned with ground-truth labels on balanced data for each dataset (300/300, 250/250/250, 250/250/250/250 samples for 2,3,4-class cases). Comparison of the first two rows reveals that the oracle choice on balanced data may not be suitable for imbalanced data, while our PCut framework, although agnostic, picks more suitable $k,\sigma$ for RBF $k$-NN.
The last row presents oracle results on RBF RMD tuned to imbalanced data. This shows that our PCut on RMD, agnostic of true labels, closely approximates the oracle performance. In addition both tables show that
our RMD graph parameterization performs consistently better than other methods. 
Similarly, Table \ref{tab:real_SSL} shows the performance of different graph constructions for SSL tasks with GRF and GTAM algorithms. We again observe that the RMD graph construction performs significantly better than all other constructions in the SSL tasks in all datasets.

\begin{table*}[htb]
\caption{Error rate performance of GRF and GTAM for imbalanced real datasets. Our method performs significantly better than other methods.}
\centering
\begin{tabular}{|c|c||c|c|c|c|c|c|c|c|c|}
  \hline
  \multicolumn{2}{|c||}{\multirow{2}{*}{Error Rates (\%)}}  &   \multicolumn{2}{c|}{USPS}  &   \multicolumn{2}{c|}{SatImg}  &   \multicolumn{3}{c|}{OptDigit}   & \multicolumn{2}{c|}{LetterRec} \\
  \cline{3-11}
  \multicolumn{2}{|c||}{}  & 8vs6 & 1,8,3,9 & 4vs3 & 1,4,7 & 6vs8 & 8vs9 & 6,1,8 & 6vs7 & 6,7,8 \\
  \hline\hline
  \multirow{4}{*}{GRF}
    & RBF $k$-NN            & 5.70 & 13.29 & 14.64 & 16.68 & 5.68  & 7.57  & 7.53 & 7.67 & 28.33 \\
    & RBF $b$-matching      & 6.02 & 13.06 & 13.89 & 16.22 & 5.95  & 7.85  & 7.92 & 7.82 & 29.21 \\
    & full-RBF              & 15.41 & 12.37 & 14.22 & 17.58 & 5.62 & 9.28 & 7.74 & 11.52 & 28.91 \\
    & full-aRBF             & 12.89 & 11.74 & 13.58 & 17.86 & 5.78 & 8.66 & 7.88 & 10.10 & 28.36 \\
    & RBF RMD               & 1.08  & 10.24 & 9.74 & 15.04 & 2.07  & 2.30  & 5.82 & 5.23 & 27.24 \\
  \hline
  \multirow{4}{*}{GTAM}
    & RBF $k$-NN            & 4.11  & 10.88 & 26.63 & 20.68 & 11.76 & 5.74  & 12.68 & 19.45 & 27.66 \\
    & RBF $b$-matching      & 3.96  & 10.83 & 27.03 & 20.83 & 12.48 & 5.65  & 12.28 & 18.85 & 28.01 \\
    & full-RBF              & 16.98  & 11.28 & 18.82 & 21.16 & 13.59 & 7.73 & 13.09 & 18.66 & 30.28 \\
    & full-aRBF             & 13.66  & 10.05 & 17.63 & 22.69 & 12.15 & 7.44 & 13.09 & 17.85 & 31.71 \\
    & RBF RMD               & 1.22  & 9.13 & 18.68 & 19.24 & 5.81  & 3.12  & 10.73 & 15.67 & 25.19 \\
  \hline
\end{tabular}
\label{tab:real_SSL}
\end{table*}

\subsection{Community detection with connectivity networks}

In this section we consider the adaptation of RMD to connectivity networks as described in Sec.~\ref{subsec:rmd_c}. We first consider performance on synthetic networks using the stochastic block model for graph generation with two clusters of sizes $n_1$ and $n_2 = n - n_1$ nodes with imbalance coefficient $\alpha = \frac{n_1}{n} \leq 0.5$. The two clusters each follow an Erd\H{o}s-R\'enyi model with edge probabilities $p_1$ and $p_2$ respectively and inter-cluster edge probabilities $q$. 

\begin{figure}[tb]
  \centering
  \includegraphics[width = .7\textwidth]{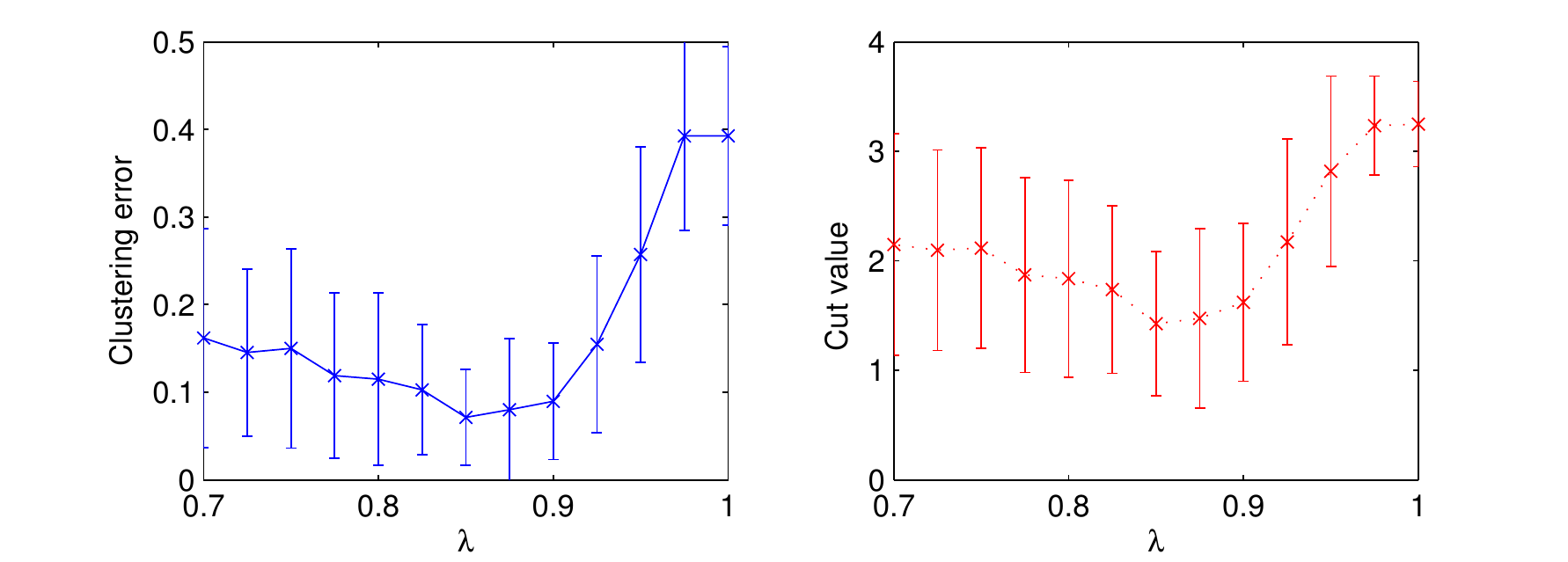}
  \caption{Average clustering error and normalized cut values with standard deviation error bars for different parameterizations $\lambda$, computed over 20 simulated graphs. The parameter that minimizes cut value also minimizes clustering error and provides an $\sim 80\%$ decrease. in clustering error on average, compared to the baseline given by SC on the original graph ($\lambda = 1$).}
  \label{fig:sbm_simple}
\end{figure}

We first consider an illustrative example in how $\lambda$ affects the cut value and clustering error with fixed imbalance $\alpha = 0.05$, $n = 500$, $p_1 = 0.2$ and $q = 0.03$ in Fig.~\ref{fig:sbm_simple} over 20 generated graphs. $p_2$ is chosen such that the expected degree of each node is equal, in order to prevent clustering using node degrees. We first observe that the cut value is a good indicator for clustering performance for different $\lambda$, i.e.~the parameter value chosen by Eq.~\eqref{eq:selection} also minimizes the clustering error (all shown cuts satisfied the size constraint with $\delta = \alpha$). We also observe that the parameter that minimizes cut value decreases the clustering error from about $40\%$ in the baseline case (which performs SC on the given graph and corresponds to $\lambda = 1$) to about $7\%$, representing an $\sim 80\%$ decrease.

\begin{figure}[tb]
  \centering
  \includegraphics[width = .7\textwidth]{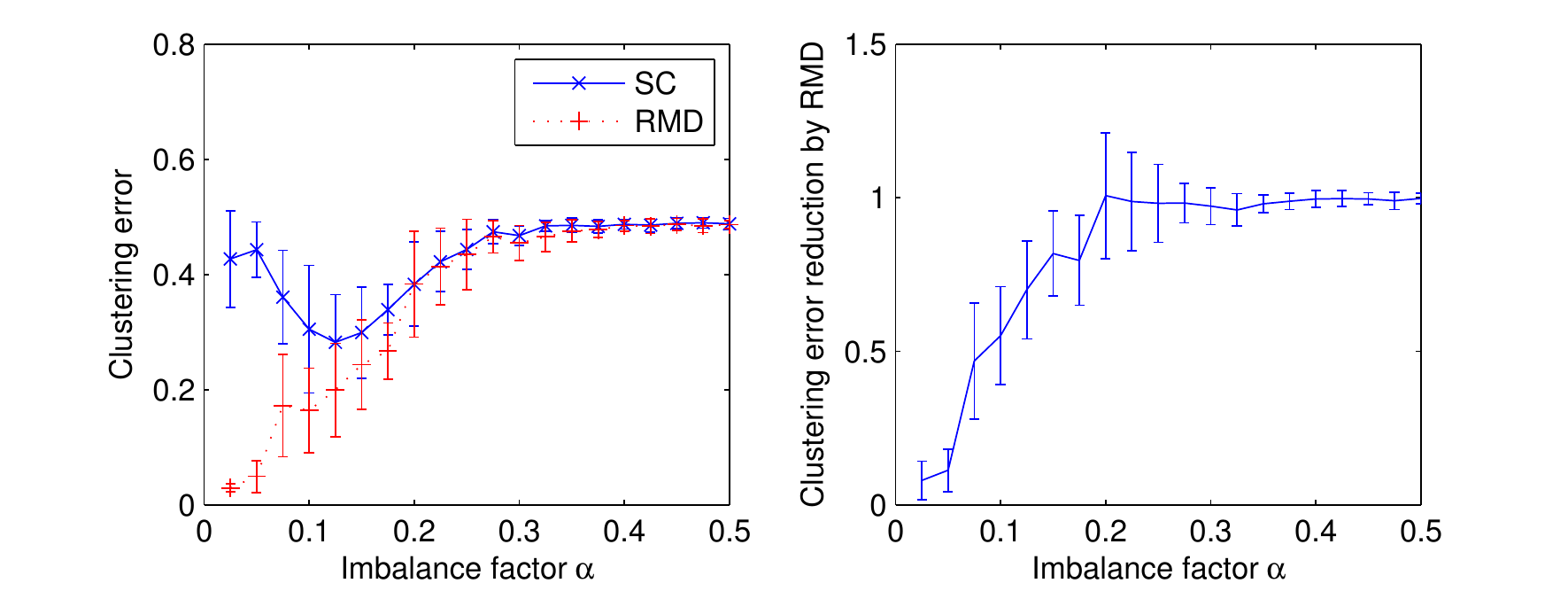}
  \caption{Average clustering error for SC and RMD with standard deviation error bars for different imbalance parameters $\alpha$ on the left figure. Right figure shows the average and standard deviation error bars for RMD error to SC error ratio. Both statistics computed over 20 simulated graphs for each $\alpha$. While RMD does not provide much gain in performance when $\alpha > 0.2$, for smaller imbalance factors we observe an up to $90\%$ reduction in clustering error.}
  \label{fig:sbm_size}
\end{figure}

Using the same graph generation model, we next investigate the effect of RMD on the performance of community detection for different imbalance coefficients and graph parameters in Fig.~\ref{fig:sbm_size}. We again consider $n = 500$ nodes with imbalance $\alpha$ varying between $0.025$ and $0.5$. We set $p_1 = 0.2$ and $q$ scaling proportionally with $1/\alpha$ which we normalize such that it is equal to $0.03$ when $\alpha = 0.05$. We choose $p_2$ such that the expected node degrees are uniform as in the previous example. To obtain the clustering with RMD we optimize $\lambda$ over the interval $[0.5, 1]$ with $0.025$ increments and choose the parameter that minimizes the cut value, as in Eq.~\eqref{eq:selection}. We observe that RMD does not provide significant performance improvements for balanced cluster sizes, however it performs significantly better compared to SC on the baseline graph for imbalanced cluster sizes $\alpha \leq 0.2$ as expected. 
We also remark that in Fig.~\ref{fig:sbm_size} the reason the error reduction factor exceeds 1 at times is the mismatch between the parameters that minimize the cut value and the clustering error (which should ideally be less than or equal to SC error, since $\lambda = 1$ is in the parameter set).

\begin{figure}[tb]
  \centering
  \includegraphics[trim={1.7cm 1.5cm 1.1cm 1.2cm},clip,width=.3\textwidth]{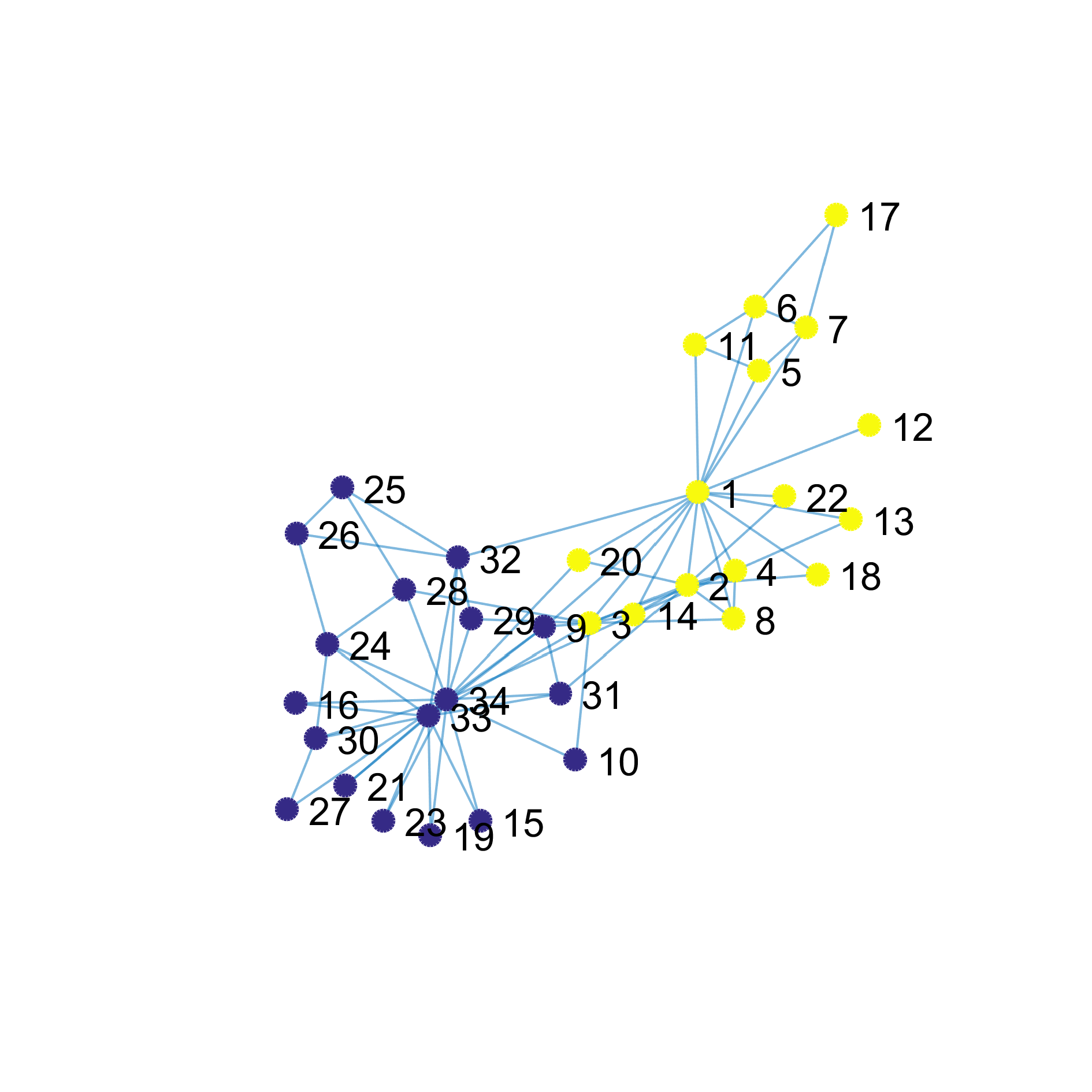}
  \includegraphics[trim={1.7cm 1.5cm 1.1cm 1.2cm},clip,width=.3\textwidth]{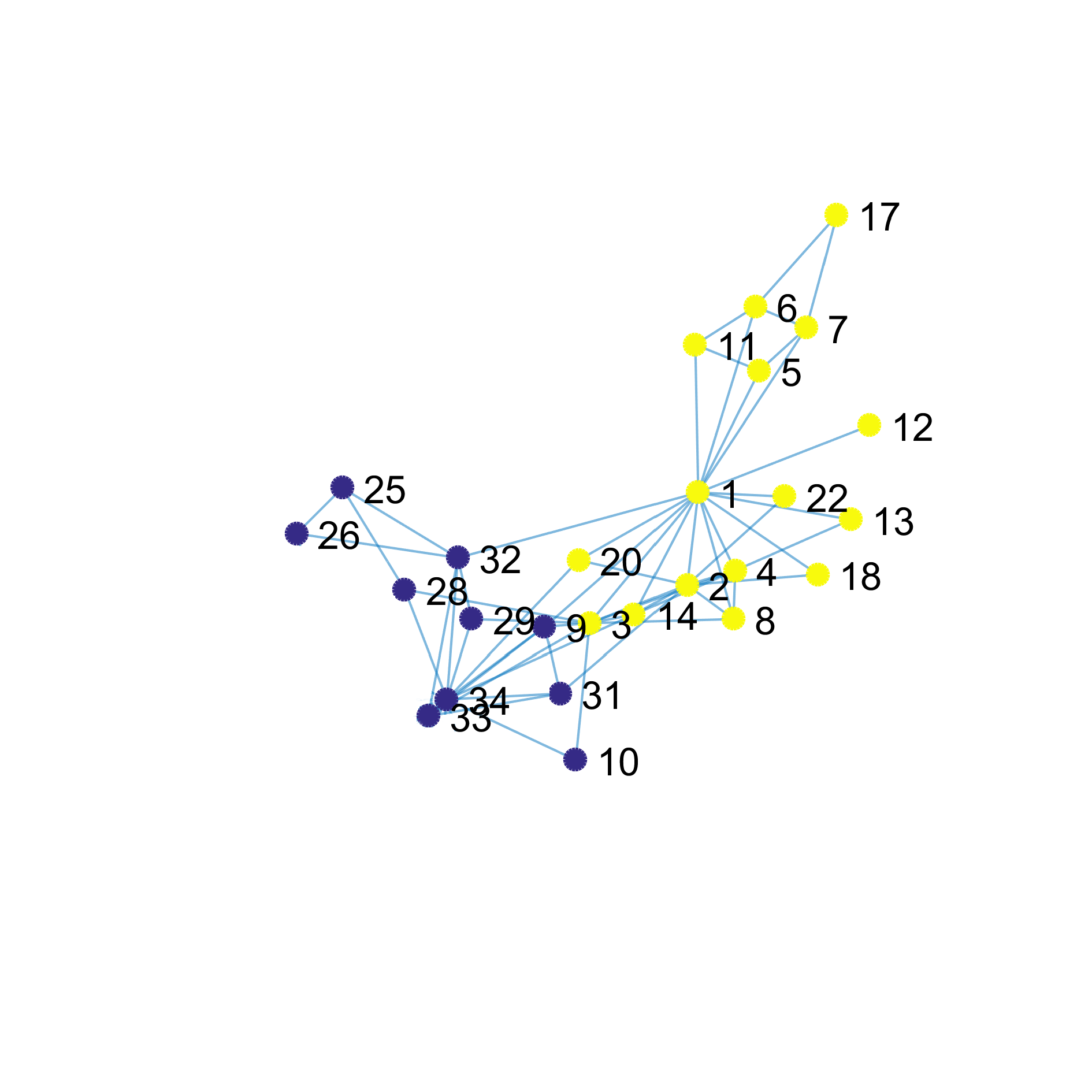}
  \caption{Visualization of Zachary's karate club social network on the left figure. Right figure illustrates the under-observed network that we consider with less balanced community sizes.}
  \label{fig:karate}
\end{figure}

\noindent
\textbf{Performance on real world examples:} \\
In this section we consider two social network datasets with well-established community structures. The first dataset we consider is the network from Zachary's karate club study \cite{zachary} which is widely used to evaluate community detection methods \cite{fortunato}. Zachary observed 34 members of a karate club over two years where the group split into two separate clubs after a disagreement. These two clubs constitute the two ground truth communities of 16 and 18 nodes, around the instructor (node 1) and administrator (node 34) of the club respectively. The network with nodes corresponding to members and edges corresponding to binary friendship indicators as determined by Zachary is illustrated in the left figure of Fig.~\ref{fig:karate}, where the coloring indicates the eventual membership of the two clubs.

Evaluating the baseline SC method and PCut using RMD on the network, we observe that only node 3 is misattributed to the wrong community resulting in good performance in both cases, which is to be expected on the relatively small and simple dataset. We then consider an under-observed and more imbalanced version of the network illustrated on the right figure of Fig.~\ref{fig:karate}, where we removed 8 outlying nodes on the blue community with node numbers 15, 16, 19, 21, 23, 24, 27 and 30. Evaluating SC on the reduced graph, we observe 10 misattributed nodes with node numbers 2, 3, 4, 8, 12, 13, 14, 18, 20 and 22, while RMD is successful in recovering all but node 3's community attribution as in the full graph using minimum cluster size parameter $\delta = 5/26$.

\begin{figure}[tb]
  \centering
  \includegraphics[width=.5\textwidth]{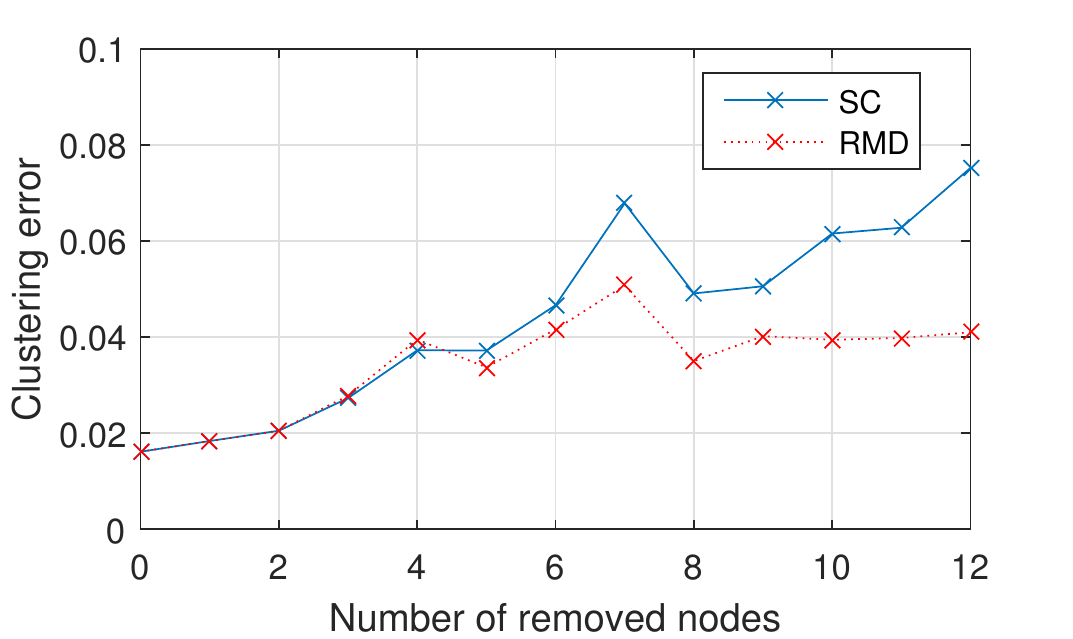}
  \caption{Average clustering error for bottlenose dolphins network with randomly removed number of nodes varying from 1 to 12. Disconnected nodes that are left after removing the initial nodes are also removed. Minimum community size parameter is selected as $\delta = 0.1$ for PCut with RMD.}
  \label{fig:dolphin}
\end{figure}

The second dataset we consider is the network of 62 bottlenose dolphins living in Doubtful Sound, New Zealand analyzed by \cite{dolphin}, which is another dataset widely used for benchmark purposes \cite{fortunato}. The edges in the network were determined by sightings of pairs of dolphins and the two ground truth communities of sizes 20 and 42 correspond to dolphins that separated after a dolphin left the area for some time. As the communities are internally well-connected with internal cliques and only six edges between the communities, SC is successful in recovering all but one node association. For a meaningful comparison, we again consider an under-observed version of the graph with nodes randomly removed from the small community. We consider 1 to 12 removed nodes with 100 different samplings each and illustrate the error rates for SC and RMD in Fig.~\ref{fig:dolphin}. We observe that as the number of removed nodes increases, the graph is more imbalanced and the error rate of both methods increase. However, PCut with RMD performs consistently better on more imbalanced cluster sizes, with up to 40\% decrease in average error rates.

Finally, we evaluate our method on experiments performed on graphs generated by the LFR benchmark algorithm \cite{lfr}. The benchmark algorithm accounts for the power-law behavior of both degree and community size distributions in real networks, resulting in more realistic network structures compared to SBM that we considered before. We refer the reader to \cite{lfr} for more details on the algorithm. While the previous two real networks we considered were small in size and had two communities, with these benchmarks we will investigate the behavior of PCut with RMD in networks with a larger number of nodes and communities. For the experiments in this section we consider 200 nodes, degree distribution parameter $\gamma = 2$, average degree $k = 10$, $k_{max} = 20$, mixing parameter $\mu = 0.1$, community size distribution parameter $\beta = 1$ and maximum community size $s_{max} = 50$. We vary the minimum community size parameter $s_{min}$ between 10 and 50 in 10 logarithmic increments and sample 100 graphs each to get a good spread of imbalanced and balanced communities, obtaining 1000 graphs in total. We note that $s_{min} = 50$ results in an exactly balanced network with 4 communities of 50 nodes each. We also remark that the number of communities is variable and random, and increases with decreased $s_{min}$, with as many as 12 communities for small $s_{min}$.  
We choose minimum partition size parameter $\delta = 10/200$ for all graphs. 
To compute the error of a given partitioning, we use the Hungarian algorithm \cite{munkres} that finds the optimal permutation of found community labels to ground truth labels by solving a minimum weighted bipartite matching problem, with assignment cost between two communities $C_i$, $C_j$ set equal to $|C_i \cup C_j| - |C_i \cap C_j|$.

\begin{figure*}[tb]
  \centering
  \begin{minipage}[t]{.32\textwidth}
    \includegraphics[width = 1\textwidth]{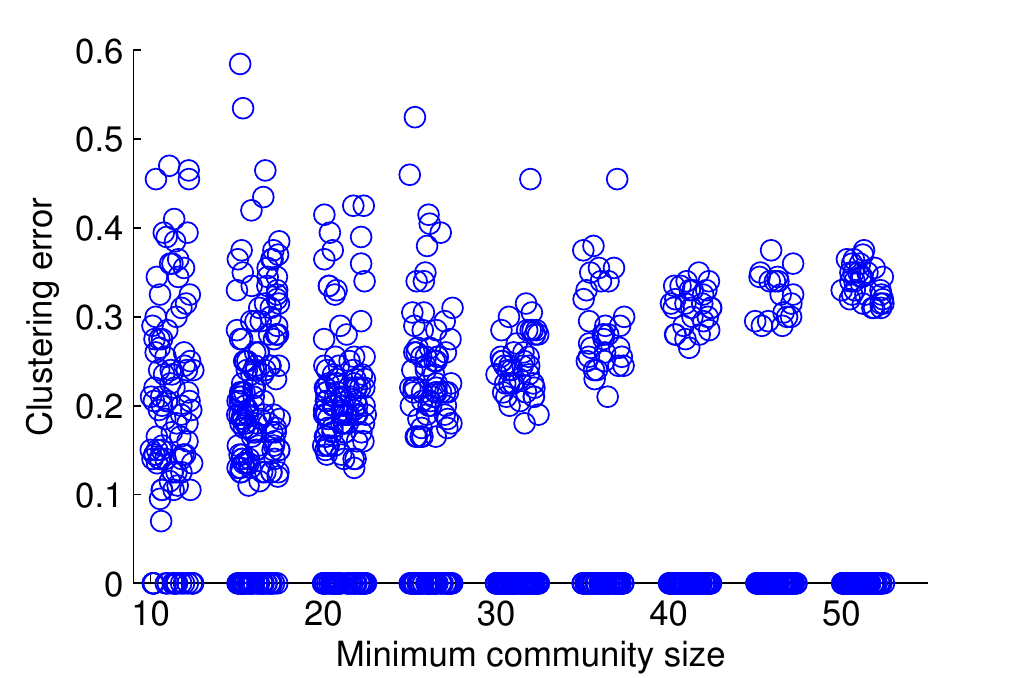}
    \makebox[6cm]{(a) SC}
  \end{minipage}
  \begin{minipage}[t]{.32\textwidth}
    \includegraphics[width = 1\textwidth]{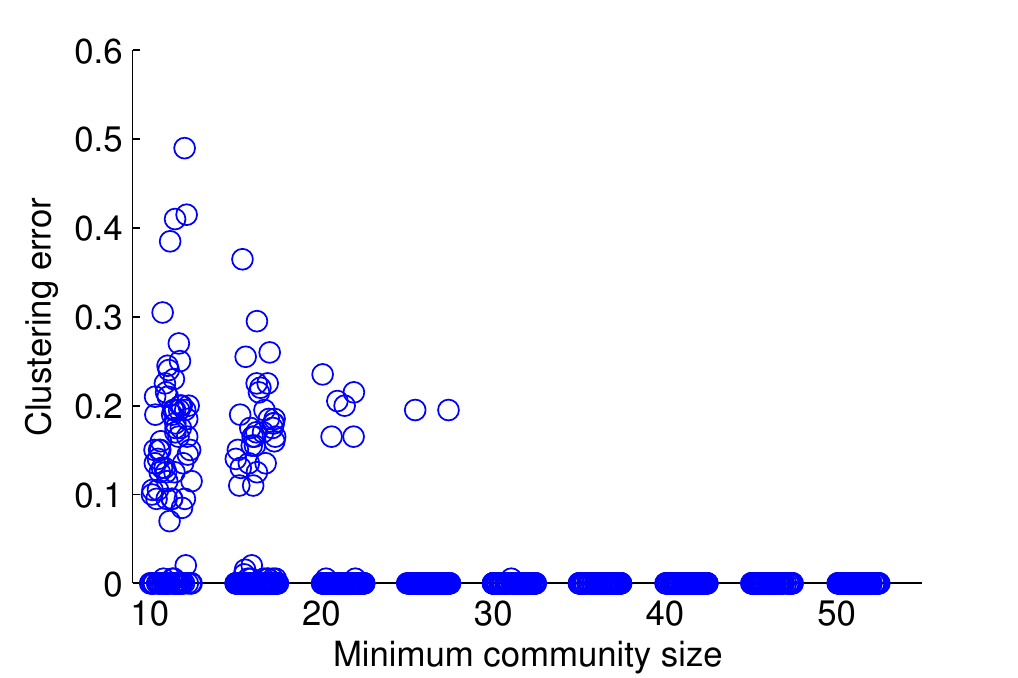}
    \makebox[6cm]{(b) PCut with RMD}
  \end{minipage}
  \begin{minipage}[t]{.34\textwidth}
    \includegraphics[trim={0cm 0.25cm 0cm 0cm},clip,width = 1.05\textwidth]{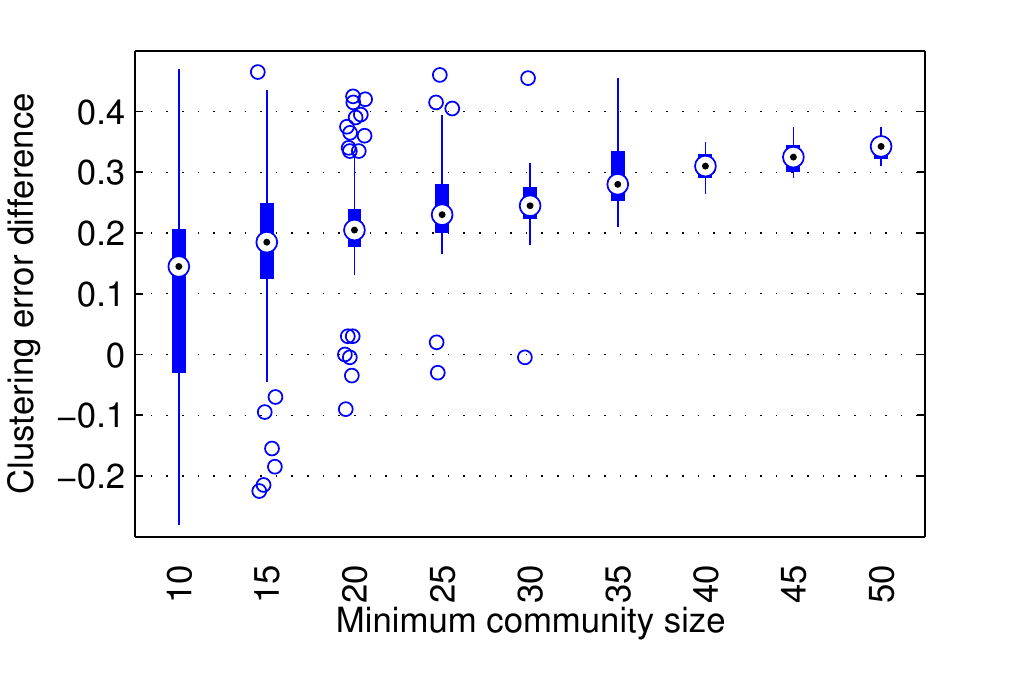}
    \makebox[6cm]{(c) Absolute improvement}
  \end{minipage}
  \caption{(a) Scatter plot of clustering error for baseline SC, (b) for PCut with RMD for 1000 generated graphs. (c) Box plot of absolute improvement in error rate from SC to RMD, ignoring samples with zero SC and RMD error. The x-axis in all graphs is the minimum community size in a given graph, while y-axis is the clustering error. In all three plots, minimum community sizes are quantized to bins with a width of 5 such that the first group in the plots corresponds to minimum community sizes of 10-14, second group of 15-19 and last group of only 50. Jittering is also added to the points on the horizontal axis to better visualize overlapping values in scatter plots (a,b).}
  \label{fig:LFR}
\end{figure*}

\begin{figure}[tb]
  \centering
  \begin{minipage}[t]{.25\textwidth}
    \includegraphics[trim={1.2cm 1cm 1.2cm 0.9cm},clip,width = 1\textwidth]{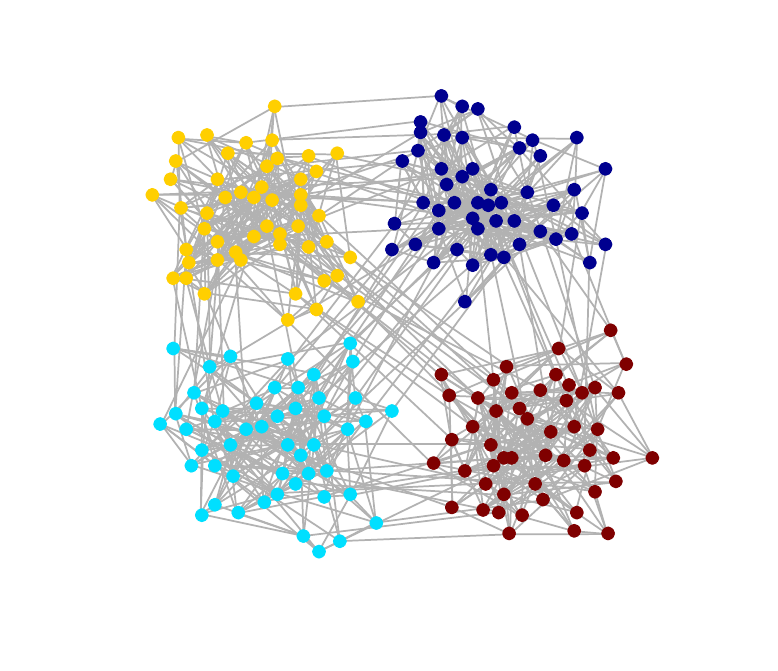}
    \makebox[5cm]{(a)}
  \end{minipage}
  \begin{minipage}[t]{.25\textwidth}
    \includegraphics[trim={1.2cm 1cm 1.2cm 0.9cm},clip,width = 1\textwidth]{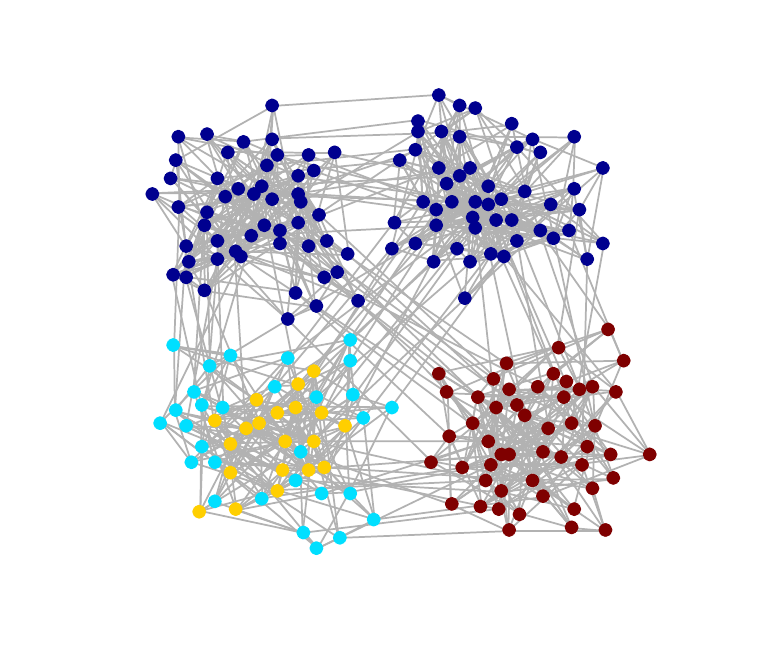}
    \makebox[5cm]{(b)}
  \end{minipage}
  \begin{minipage}[t]{.25\textwidth}
    \includegraphics[trim={1.2cm 1cm 1.2cm 0.9cm},clip,width = 1\textwidth]{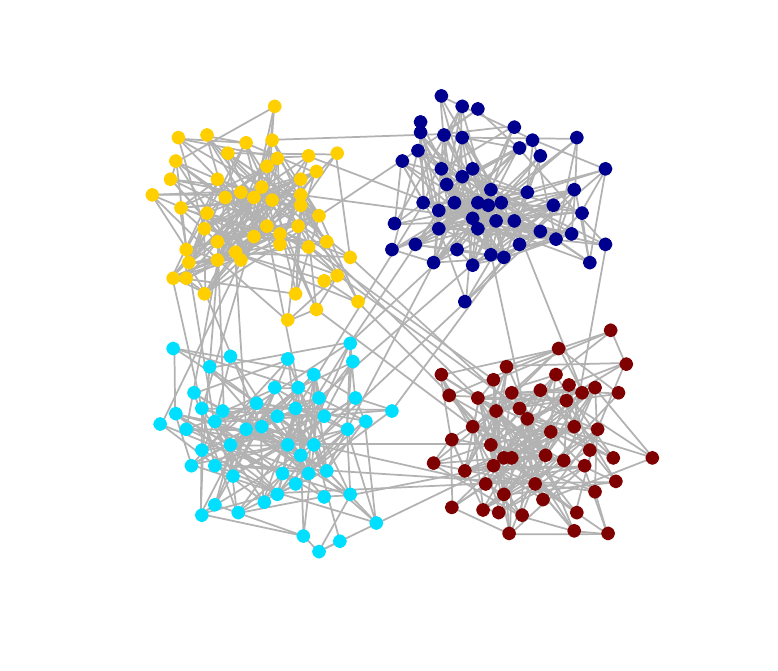}
    \makebox[5cm]{(c)}
  \end{minipage}
  \caption{Example network generated by LFR with exactly balanced communities. (a) Ground truth communities, (b) SC estimation, (c) PCut estimation on RMD graph with $\lambda = 0.75$. RMD parameterization emphasizes the separation between communities and prevents spectral clustering from merging the two communities on the top.}
  \label{fig:LFR_sample}
\end{figure}

Using the minimum community size in a generated graph as an indicator for imbalancedness of community sizes, we plot clustering error vs.\ minimum community size for SC and PCut with RMD in Fig.~\ref{fig:LFR}. We observe a distinct bimodal behavior in the error rates for SC, where there is a significant number of samples with approximately zero error (the fraction of which increases with the minimum community size) and the rest of the samples follow a roughly linearly increasing error rate. This error behavior is due to the algorithm merging separate clusters and separating a cluster to multiple clusters in certain cases, as consistent with the error rates being near the cluster sizes for the balanced case (minimum size close to 50). We illustrate this with one of the generated balanced networks with $35\%$ SC error in Fig.~\ref{fig:LFR_sample}. We see that SC has merged the top two clusters together while separating the bottom-left cluster to two, causing the high error. However, the optimal PCut corresponding to RMD graph with $\lambda = 0.75$ has emphasized the separation of the four communities, leading to the correct assignment of nodes. While RMD graphs provide error improvement for imbalanced clusters as expected, we see that it can solve the cluster merging and separating problems by emphasizing the separation between clusters, reducing the error rate to zero.
We can observe from Fig.~\ref{fig:LFR} that PCut improves upon SC with a $20\%$ median absolute improvement in the error rate for imbalanced graphs, to up to $35\%$ median absolute improvement in balanced cases.

\section{Conclusion}
\label{sec:conclusion}

In this paper we investigated the performance of spectral clustering based on minimizing RCut or NCut objectives for graph partitioning with imbalanced partition sizes and showed that these objectives may lead to poor clustering performance in both similarity and connectivity network modalities. To this end we proposed the partition constrained min-cut (PCut) framework, which seeks min-cut partitions under minimum cluster size constraints. Since constrained min-cut is NP-hard, we adopt existing spectral methods (SC, GRF, GTAM) as a black-box subroutine on a parameterized family of graphs to generate candidate partitions and solve PCut on these partitions. We proposed rank modulated degree (RMD) graphs as a rich graph parameterization based on adaptively modulating the node degrees in varying levels to adapt to different levels of imbalanced data.
Our framework automatically selects the parameters based on PCut objective, and can be used in conjunction with other graph-based partition methods such as 1-spectral clustering, Cheeger cut or sparsest cut \cite{Buhler09,Hein10,Szlam10,Arora09}.
Our idea is justified through limit cut analysis and both synthetic and real experiments on clustering and SSL tasks for the similarity networks, and community detection for connectivity networks.

\paragraph*{Acknowledgments} This material is based upon work supported in part by the U.S.
Department of Homeland Security, Science and Technology
Directorate, Office of University Programs, under Grant Award 2013-ST-061-ED0001, the National Science Foundation under Grants CCF-1320547 and 1218992, and by ONR Grant 50202168. The views and conclusions
contained in this document are those of the authors and
should not be interpreted as necessarily representing the
official policies, either expressed or implied, of the U.S.
Department of Homeland Security, NSF and ONR.

\section{Appendix: Proofs of Theorems}
For convenience in analysis, let $n=m_1(m_2+1)$ and divide $n$ data points into $m_1 + 1$ sets $D=D_0 \bigcup  D_1 \bigcup ... \bigcup D_{m_1}$, where $D_0=\{x_1,...,x_{m_1}\}$ and each $D_j$, $j=1,...,m_1$ contains $m_2$ points. $D_j$ is used to generate the statistic $\eta$ for $u$ and $x_j \in D_0$ for $j=1,...,m_1$. $D_0$ is used to compute the rank of $u$ with the formula
\begin{equation}
    R(u) = \frac{1}{m_1}\sum_{j=1}^{m_1} \mathbb{I}_{\{ \eta(x_j;D_j)>\eta(u;D_j) \}}.
\end{equation}
We analyze and prove our result for the statistic $\eta(u)$ of the form
\begin{equation}
  \eta(u;D_j) = \frac{1}{l}\sum^{l+\lfloor \frac{l}{2} \rfloor}_{i=l-\lfloor \frac{l-1}{2} \rfloor}\left( \frac{l}{i} \right)^{\frac{1}{d}}D_{(i)}(u),
\end{equation}
where we used $l$ in place of $k_0$ and $D_{(i)}(u)$ denotes the distance from $u$ to its $i$-th nearest neighbor among $m_2$ points in $D_j$. In practice we can omit the weight and use the average of first to $l$-th nearest neighbor distances as described in Sec.~\ref{sec:RMD_idea}.

\vspace{5pt}
\noindent\textbf{Proof of Theorem \ref{rank-pvalue}:}

The proof involves two steps:
\begin{itemize}
  \item[1.] The expectation of the empirical rank $\mathbb{E}\left[R(u)\right]$ is shown to converge to $p(u)$ as $n\rightarrow\infty$.
  \item[2.] The empirical rank $R(u)$ is shown to concentrate at its expectation as $n\rightarrow\infty$.
\end{itemize}
The first step is shown through Lemma \ref{lem:expectation}. For the second step, notice that the rank $R(u) = \frac{1}{m_1}\sum_{j=1}^{m_1} Y_j$, where $Y_j = \mathbb{I}_{\{ \eta(x_j;D_j)>\eta(u;D_j) \}}$ is independent across different $j$'s, and $Y_j \in [0,1]$. By Hoeffding's inequality, we have
\begin{equation}
    \mathbb{P}\left( | R(u) - \mathbb{E}\left[R(u)\right] | > \epsilon \right) < 2\exp\left( -2m_1\epsilon^2 \right).
\end{equation}
Combining these two steps completes the proof.

\vspace{5pt}
\noindent\textbf{Proof of Theorem \ref{part2}:}

We want to establish convergence results for the cut term and the balancing terms respectively, that is:
\begin{align}
  \frac{1}{nk_n}\sqrt[d]{\frac{n}{k_n}}cut_n(S)
    & \longrightarrow C_d\int_S{f^{1-\frac{1}{d}}(s)\rho(s)^{1+\frac{1}{d}} \dx s} \label{eq:term1}\\
  n \frac{1}{|V^\pm|} & \longrightarrow \frac{1}{\mu(C^\pm)} \label{eq:term2R}  \\
  n k_n\frac{1}{vol(V^\pm)} & \longrightarrow \frac{1}{\mu(C^\pm)}, \label{eq:term2N}
\end{align}
where $V^\pm = \{x \in V: x \in C^\pm\}$ are the discrete versions of $C^\pm$.

The balancing terms Eq.~\eqref{eq:term2R}, \eqref{eq:term2N} are obtained similarly using Chernoff bound on the sum of binomial random variables, since the number of points in $V^\pm$ is binomially distributed $Binom(n,\mu(C^\pm))$. Details can be found in \cite{Maier1}.

Eq.~\eqref{eq:term1} is established in two steps. First we can show that the LHS cut term converges to its expectation $\mathbb{E}\left(\frac{1}{nk_n}\sqrt[d]{\frac{n}{k_n}}cut_n(S)\right)$ by McDiarmid's inequality. Second we show this expectation term actually converges to the RHS of Eq.~\eqref{eq:term1}. This is shown in Lemma~\ref{expectation}.

\begin{lemma}\label{expectation}
Given the assumptions of Theorem 2,
\begin{equation*}
  \mathbb{E}\left(\frac{1}{nk_n}\sqrt[d]{\frac{n}{k_n}}cut_n(S)\right) \longrightarrow C_d\int_S{f^{1-\frac{1}{d}}(s)\rho(s)^{1+\frac{1}{d}} \dx s}.
\end{equation*}
where $C_d=\frac{2\eta_{d-1}}{(d+1)\eta_d^{1+\frac{1}{d}}}$.
\end{lemma}

\begin{IEEEproof}
The proof is an extension of \cite{Maier2}. For simplicity of exposition we provide an outline of the extension and further details can be found in \cite{Maier1}. The first trick is to define a cut function for a fixed point $x_i\in V^+$ whose expectation is easier to compute,
\begin{equation}
cut_{x_i} = \sum_{v\in V^{-},(x_i,v)\in E}w(x_i,v).
\end{equation}
$cut_{x_i}$ is defined similarly for $x_i\in V^-$. The expectation of $cut_{x_i}$ and $cut_n(S)$ can be shown to satify
\begin{eqnarray}\label{eq:expect}
\mathbb{E}(cut_n(S))=n\mathbb{E}_x(\mathbb{E}(cut_{x})).
\end{eqnarray}
Then the value of $\mathbb{E}(cut_{x_i})$ can be computed as
\begin{equation}
  (n-1) \int_0^{\infty} \left[ \int_{B(x_i,r)\cap{C^-}} f(y) \dx y\right] \dx F_{R_{x_i}^k}(r),
\end{equation}
where $r$ is the distance of $x_i$ to its $k_n\rho(x_i)$-th nearest neighbor. The value of $r$ is a random variable and can be characterized by the CDF $F_{R_{x_i}^k}(r)$.
Combining above with Eq.~\eqref{eq:expect} we can write down the whole expected cut value
\begin{align}
  \mathbb{E}(cut_n(S)) = n \mathbb{E}_x(\mathbb{E}(cut_{x})) = n \int_{\mathbb{R}^d}f(x) \mathbb{E}(cut_{x}) \dx x = n(n-1) \int_{\mathbb{R}^d} f(x) \left[ \int_0^{\infty}{g(x,r) \dx F_{R_x^k}(r)} \right] \dx x,
\end{align}
where to simplify the expression we used $g(x,r)$ to denote
\begin{equation}
    g(x,r)=\begin{cases}
               \int_{B(x,r)\cap{C^-}}f(y)dy,  x\in{C^+} \\
               \int_{B(x,r)\cap{C^+}}f(y)dy,  x\in{C^-}.
             \end{cases}
\end{equation}

Under general assumptions, the random variable $r$ will highly concentrate around its mean $\mathbb{E}(r_x^k)$ when $n$ tends to infinity.
Furthermore, as $k_n/n\rightarrow{0}$, $\mathbb{E}(r_x^k)$ tends to zero and the speed of convergence is given by
\begin{equation}\label{eq:EkNN}
  \mathbb{E}(r_x^k) \approx \left( \frac{k \rho(x)}{(n-1) f(x) \eta_d} \right)^{\frac{1}{d}}.
\end{equation}
Therefore the inner integral in the cut value can be approximated by $g(x,\mathbb{E}(r_x^k))$, which implies that
\begin{equation}
    \mathbb{E}(cut_n(S))\approx{n}(n-1)\int_{\mathbb{R}^d}f(x)g(x,\mathbb{E}(r_x^k)) \dx x.
\end{equation}

The next trick is to decompose the integral over $\mathbb{R}^d$ into two orthogonal directions, i.e.\ the direction along the hyperplane $S$ and its normal direction
\begin{align*}
  \int_{\mathbb{R}^d} & f(x) g(x,\mathbb{E}(r_x^k)) \dx x = \int_{S}\int_{-\infty}^{+\infty}f(s+t\vec{n})g(s+t\vec{n},\mathbb{E}(r_{s+t\vec{n}}^k)) \dx t \dx s,
\end{align*}
where we used $\vec{n}$ to denote the unit normal vector.
When $t>\mathbb{E}(r_{s+t\vec{n}}^k)$, the integral region of $g$ will be empty, i.e.\ $B(x,\mathbb{E}(r_x^k))\cap{C^-}=\emptyset$. On the other hand, when $x=s+t\vec{n}$ is close to $s\in{S}$, we have the approximation $f(x)\approx{f(s)}$
\begin{align*}
  \int_{-\infty}^{+\infty} f(s+t\vec{n})g(s+t\vec{n},\mathbb{E}(r_{s+t\vec{n}}^k)) \dx t & \approx 2\int_{0}^{\mathbb{E}(r_{s}^k)}f(s)\left[f(s)vol\left(B(s+t\vec{n},\mathbb{E}{r_s^k})\cap{C^-}\right)\right] \dx t  \\
  & = 2f^2(s)\int_{0}^{\mathbb{E}(r_{s}^k)}vol\left(B(s+t\vec{n},\mathbb{E}(r_s^k))\cap{C^-}\right) \dx t.
\end{align*}

The term $vol\left(B(s+t\vec{n},\mathbb{E}(r_s^k))\cap{C^-}\right)$ is the volume of the $d$-dimensional spherical cap of radius $\mathbb{E}(r_s^k))$, which is at distance $t$ to the center. Through direct computation we obtain
\begin{equation*}
  \int_{0}^{\mathbb{E}(r_{s}^k)}vol\left(B(s+t\vec{n},\mathbb{E}(r_s^k))\cap{C^-}\right) \dx t=\mathbb{E}(r_s^k)^{d+1}\frac{\eta_{d-1}}{d+1}.
\end{equation*}
Combining the above step and plugging in the approximation of $\mathbb{E}(r_s^k)$ in Eq.~\eqref{eq:EkNN}, we finish the proof.
\end{IEEEproof}

\begin{lemma}\label{lem:expectation}
By choosing $l$ properly, it follows that as $m_2\rightarrow\infty$,
\[ | \mathbb{E}\left[R(u)\right] - p(u)| \longrightarrow 0. \]
\end{lemma}

\begin{IEEEproof}
Taking the expectation with respect to $D$ we have
\begin{align*}
  \mathbb{E}_D\left[R(u)\right] & = \mathbb{E}_{D\backslash D_0} \left[\mathbb{E}_{D_0} \left[ \frac{1}{m_1} \sum_{j=1}^{m_1} \mathbb{I}_{\{\eta(u;D_j)<\eta(x_j;D_j)\}} \right] \right]\\
  & = \frac{1}{m_1} \sum_{j=1}^{m_1} \mathbb{E}_{x_j} \left[ \mathbb{E}_{D_j}\left[
\mathbb{I}_{\{\eta(u;D_j)<\eta(x_j;D_j)\}} \right] \right] \\
  & = \mathbb{E}_x \left[ \mathcal{P}_{D_1} \left( \eta(u;D_1)<\eta(x;D_1) \right) \right]
\end{align*}
The last equality holds due to the i.i.d.\ symmetry of $\{x_1,...,x_{m_1}\}$ and $D_1,...,D_{m_1}$. We fix both $u$ and $x$ and temporarily disregard $\mathbb{E}_{D_1}$. Let $F_x(y_1,...,y_{m_2})=\eta(x)-\eta(u)$, where $y_1,...,y_{m_2}$ are the $m_2$ points in $D_1$. It follows that:
\begin{align*}
  \mathcal{P}_{D_1}\left(\eta(u)<\eta(x)\right) & =\mathcal{P}_{D_1}\left(F_x(y_1,...,y_{m_2})>0\right) = \mathcal{P}_{D_1}\left(F_x-\mathbb{E}F_x>-\mathbb{E}F_x\right).
\end{align*}

To check McDiarmid's requirements, we replace $y_j$ with $y_j'$. It is easily verified that $\forall j=1,...,m_2$,
\begin{equation*}
  |F_x(y_1,...,y_{m_2})-F_x(y_1,...,y_j',...,y_{m_2})| \leq \frac{2^{1+\frac{1}{d}} C}{l} \leq \frac{4C}{l},
\end{equation*}
where $C$ is the diameter of support. Notice despite the fact that $y_1,...,y_{m_2}$ are random vectors we can still apply McDiarmid's inequality, because due to the specific form of $\eta$, $F_x(y_1,...,y_{m_2})$ is a function of $m_2$ i.i.d random variables $r_1,...,r_{m_2}$ where $r_i$ is the distance from $x$ to $y_i$.
Therefore if $\mathbb{E}F_x<0$, or $\mathbb{E}\eta(x)<\mathbb{E}\eta(u)$, we have by McDiarmid's inequality,
\begin{align*}
    \mathcal{P}_{D_1}\left(\eta(u)<\eta(x)\right) & = \mathcal{P}_{D_1}\left( F_x > 0 \right) = \mathcal{P}_{D_1}\left( F_x-\mathbb{E}F_x>-\mathbb{E}F_x \right) \leq \exp\left(-\frac{(\mathbb{E}F_x)^2 l^2}{8C^2m_2}\right).
\end{align*}
and we can rewrite the above inequality as
\begin{equation}\label{equ:bound_no_expectation}
    \mathbb{I}_{\{\mathbb{E}F_x>0\}}-e^{-\frac{(\mathbb{E}F_x)^2 l^2}{8C^2m_2}}
    \leq \mathcal{P}_{D_1}\left( F_x > 0 \right)
    \leq \mathbb{I}_{\{\mathbb{E}F_x>0\}}+e^{-\frac{(\mathbb{E}F_x)^2 l^2}{8C^2m_2}}
\end{equation}
It can be shown that the same inequality holds for $\mathbb{E}F_x>0$, or $\mathbb{E}\eta(x)>\mathbb{E}\eta(u)$. We then take the expectation with respect to $x$
\begin{align}\label{equ:bound_with_expectation}
    \mathcal{P}_x & \left(\mathbb{E}F_x>0\right)-\mathbb{E}_x\left[e^{-\frac{(\mathbb{E}F_x)^2 l^2}{8C^2m_2}}\right] \leq \mathbb{E}\left[\mathcal{P}_{D_1}\left( F_x > 0 \right)\right] \leq \mathcal{P}_x\left(\mathbb{E}F_x>0\right)+\mathbb{E}_x\left[e^{-\frac{(\mathbb{E}F_x)^2 l^2}{8C^2m_2}}\right].
\end{align}
Divide the support of $x$ into two parts, $\mathbb{X}_1$ and $\mathbb{X}_2$, where $\mathbb{X}_1$ contains those $x$ whose density $f(x)$ is relatively far away from $f(u)$, and $\mathbb{X}_2$ contains those $x$ whose density is close to $f(u)$. We show for $x \in \mathbb{X}_1$, the above exponential term converges to 0 and $\mathcal{P}\left(\mathbb{E}F_x>0\right) = \mathcal{P}_x\left( f(u)>f(x) \right)$, while the rest $x\in\mathbb{X}_2$ has very small measure. Let $A(x)=\left(\frac{k}{f(x) c_d m_2}\right)^{\frac{1}{d}}$. By Lemma \ref{lem:bound_expectation} we have
\begin{align*}
    | \mathbb{E}\eta(x) - A(x) | & \leq \gamma \left(\frac{l}{m_2}\right)^{\frac{1}{d}} A(x) \leq \gamma \left(\frac{l}{m_2}\right)^{\frac{1}{d}} \left(\frac{l}{f_{min}c_d m_2}\right)^{\frac{1}{d}} = \left(\frac{\gamma_1}{c_d^{\frac{1}{d}}}\right) \left(\frac{l}{m_2}\right)^{\frac{2}{d}},
\end{align*}
where $\gamma$ is a constant and $\gamma_1 = \gamma \left(\frac{1}{f_{min}}\right)^{\frac{1}{d}}$. Applying the triangle inequality we have
\begin{align*}
  A(x)- & A(u)- 2\left(\frac{\gamma_1}{c_d^{\frac{1}{d}}}\right) \left(\frac{l}{m_2}\right)^{\frac{2}{d}}
  \leq \mathbb{E}\left[\eta(x) - \eta(u)\right] \leq A(x)-A(u)+ 2\left(\frac{\gamma_1}{c_d^{\frac{1}{d}}}\right) \left(\frac{l}{m_2}\right)^{\frac{2}{d}}.
\end{align*}
Now let $\mathbb{X}_1 = \left\{ x:|f(x)-f(u)| \geq 3\gamma_1 d f_{min}^{\frac{d+1}{d}} \left(\frac{l}{m_2}\right)^{\frac{1}{d}} \right\}$. For $x\in \mathbb{X}_1$, it can be verified that $|A(x)-A(u)|\geq 3\left(\frac{\gamma_1}{c_d^{\frac{1}{d}}}\right) \left(\frac{l}{m_2}\right)^{\frac{2}{d}}$, or $|\mathbb{E}\left[\eta(x) - \eta(u)\right]| > \left(\frac{\gamma_1}{c_d^{\frac{1}{d}}}\right) \left(\frac{l}{m_2}\right)^{\frac{2}{d}}$, and $\mathbb{I}_{\{f(u)>f(x)\}}=\mathbb{I}_{\{\mathbb{E}\eta(x)>\mathbb{E}\eta(u)\}}$. For the exponential term in Eq.~\eqref{equ:bound_no_expectation} we have
\begin{equation}
  \exp\left(-\frac{(\mathbb{E}F_x)^2 l^2}{2C^2m_2}\right)
  \leq \exp\left(-\frac{ \gamma_1^2 l^{2+\frac{4}{d}} }{ 8C^2 c_d^{\frac{2}{d}} m_2^{1+\frac{4}{d}} } \right).
\end{equation}
For $x \in \mathbb{X}_2 = \left\{ x : |f(x)-f(u)| < 3\gamma_1 d \left(\frac{l}{m_2}\right)^{\frac{1}{d}}f_{min}^{\frac{d+1}{d}} \right\}$, by the regularity assumption, we have $\mathcal{P}(\mathbb{X}_2)<3M\gamma_1 d \left(\frac{l}{m_2}\right)^{\frac{1}{d}}f_{min}^{\frac{d+1}{d}}$. Combining the two cases into Eq.~\eqref{equ:bound_with_expectation} we have the upper bound
\begin{align*}
  \mathbb{E}_D\left[R(u)\right] & = \mathbb{E}_x\left[\mathcal{P}_{D_1}\left(\eta(u)<\eta(x)\right)\right] \\
  & = \int_{\mathbb{X}_1}\mathcal{P}_{D_1}\left(\eta(u)<\eta(x)\right)f(x) \dx x + \int_{\mathbb{X}_2}\mathcal{P}_{D_1}\left(\eta(u)<\eta(x)\right)f(x) \dx x \\
  & \leq \left( \mathcal{P}_x\left(f(u)>f(x)\right) + \exp\left(-\frac{ \gamma_1^2 l^{2+\frac{4}{d}} }{ 8C^2 c_d^{\frac{1}{d}} m_2^{1+\frac{4}{d}} } \right) \right) \mathcal{P}(x\in \mathbb{X}_1) + \mathcal{P}(x\in \mathbb{X}_2) \\
  & \leq  \mathcal{P}_x\left(f(u)>f(x)\right) + \exp\left(-\frac{ \gamma_1^2 l^{2+\frac{4}{d}} }{ 8C^2 c_d^{\frac{1}{d}} m_2^{1+\frac{4}{d}} } \right) + 3M\gamma_1 d f_{min}^{\frac{d+1}{d}} \left(\frac{l}{m_2}\right)^{\frac{1}{d}}.
\end{align*}
Let $l=m_2^\alpha$ such that $\frac{d+4}{2d+4}<\alpha<1$, and the latter two terms will converge to 0 as $m_2 \rightarrow \infty$. The analysis for the lower bound follows along similar lines.
\end{IEEEproof}

\begin{lemma}\label{lem:bound_expectation}
  Let $A(x)=\left(\frac{l}{m c_d f(x)}\right)^{\frac{1}{d}}$, $\lambda_1 = \frac{\lambda}{f_{min}}\left(\frac{1.5}{c_d f_{min}}\right)^{\frac{1}{d}}$. For an appropriate $l$, the expectation of $l$-NN distance $\mathbb{E}D_{(l)}(x)$ among $m$ points satisfies:
\begin{equation}
  | \mathbb{E}D_{(l)}(x) - A(x) | = O\left( A(x) \lambda_1 \left(\frac{l}{m}\right)^{\frac{1}{d}} \right).
\end{equation}
\end{lemma}

\begin{IEEEproof}
Denote $r(x,\alpha)=\min\{r:\mathcal{P}\left(B(x,r)\right)\geq \alpha\}$. Let $\delta_m \rightarrow 0$ as $m \rightarrow \infty$, and $0<\delta_{m}<1/2$.
Let $U\sim Binom(m,(1+\delta_m)\frac{l}{m})$ be a binomial random variable, with $\mathbb{E}U = (1+\delta_{m})l$. We then have
\begin{align}
  \mathcal{P} & \left(D_{(l)}(x) > r\left(x,(1+\delta_m)\frac{l}{m}\right) \right) = \mathcal{P}\left(U<l\right) \\
  & = \mathcal{P}\left(U<\left(1-\frac{\delta_m}{1+\delta_m}\right)(1+\delta_m)l\right) \\
  & \leq \exp\left(-\frac{\delta_m^2 l}{2(1+\delta_m)}\right),
\end{align}
where the last inequality holds from Chernoff's bound. Letting $r_1 = r(x,(1+\delta_m)\frac{l}{m})$, $\mathbb{E}D_{(l)}(x)$ can be bounded as
\begin{align}
  \mathbb{E}D_{(l)}(x)
  & \leq r_1\left[1-\mathcal{P}\left(D_{(l)}(x)>r_1\right)\right] + C\mathcal{P}\left(D_{(l)}(x)>r_1\right)  \\
  & \leq r_1 + C \exp\left(-\frac{\delta_m^2 l}{2(1+\delta_m)}\right),
\end{align}
where $C$ is the diameter of support. Similarly we can show the lower bound
\begin{equation}
    \mathbb{E}D_{(l)}(x) \geq r(x,(1-\delta_m)\frac{l}{m}) - C \exp\left(-\frac{\delta_m^2 l}{2(1-\delta_m)}\right).
\end{equation}

We next relate $r_1$ with $A(x)$ in the upper bound. Notice that $\mathcal{P}\left(B(x,r_1)\right)=(1+\delta_m)\frac{l}{m} \geq c_d r_1^d f_{min}$, so a fixed but loose upper bound is $r_1 \leq \left(\frac{(1+\delta_m)l}{c_d f_{min} m}\right)^{\frac{1}{d}} = r_{max}$. Assume $l/m$ is sufficiently small so that $r_1$ is sufficiently small. By the smoothness condition, the density within $B(x,r_1)$ is lower-bounded by $f(x)-\lambda r_1$, so we have
\begin{align*}
  \mathcal{P}\left(B(x,r_1)\right) & = (1+\delta_m)\frac{l}{m} \geq c_d r_1^d \left( f(x)-\lambda r_1 \right) = c_d r_1^d f(x)\left( 1-\frac{\lambda}{f(x)}r_1 \right) \geq c_d r_1^d f(x)\left( 1-\frac{\lambda}{f_{min}}r_{max} \right),
\end{align*}
that is,
\begin{equation}
  r_1 \leq A(x)\left( \frac{1+\delta_m}{1-\frac{\lambda}{f_{min}}r_{max}} \right)^{\frac{1}{d}}.
\end{equation}
Inserting the expression $r_{max}$ and setting $\lambda_1 = \frac{\lambda}{f_{min}}\left(\frac{1.5}{c_d f_{min}}\right)^{\frac{1}{d}}$, we have
\begin{align*}
  \mathbb{E} D_{(l)}(x)-A(x) & \leq A(x)\left( \left(\frac{1+\delta_m}{1-\lambda_1 \left(\frac{l}{m}\right)^{\frac{1}{d}}}\right)^{\frac{1}{d}} \!\!\! -1 \right) + C \exp\left(-\frac{\delta_m^2 l}{2(1+\delta_m)}\right) \\
  & \leq A(x)\left( \frac{1+\delta_m}{1-\lambda_1 \left(\frac{l}{m}\right)^{\frac{1}{d}}}-1 \right)  + C \exp\left(-\frac{\delta_m^2 l}{2(1+\delta_m)}\right) \\
  & = A(x)\frac{\delta_m + \lambda_1 \left(\frac{l}{m}\right)^{\frac{1}{d}}}{1-\lambda_1\left(\frac{l}{m}\right)^{\frac{1}{d}}} + C \exp\left(-\frac{\delta_m^2 l}{2(1+\delta_m)}\right) \\
  & = O\left( A(x) \lambda_1 \left(\frac{l}{m}\right)^{\frac{1}{d}} \right).
\end{align*}
The last equality follows by letting $l=m^{\frac{3d+8}{4d+8}}$ and $\delta_m=m^{-\frac{1}{4}}$. The lower bound follows along similar lines.

\end{IEEEproof}

\bibliographystyle{IEEEtran}
\bibliography{RMD_bib}

\end{document}